\documentclass[journal]{IEEEtran}
\usepackage{graphicx}
\usepackage{multirow}
\usepackage{epstopdf}
\usepackage{times}
\usepackage{amsmath}
\usepackage{amssymb}
\usepackage{amsfonts}
\usepackage{color}
\usepackage{subfigure}
\usepackage{hyperref}
\usepackage{flushend}

\usepackage{algorithm}
\usepackage{algorithmic}
\usepackage{rotating}

\usepackage{adjustbox,lipsum}
\usepackage{dsfont}
\usepackage{booktabs}
\usepackage{multirow}
\usepackage{array}

\usepackage{pifont}

\newcommand{\xmark}{\ding{55}}%

\newcommand{\etal}{{\em et al.\,}}       
\newcommand{\eg}{{\em e.g.}}           
\newcommand{\ie}{{\em i.e.}}           

\begin{document}

\title{A Novel Mix-normalization Method for Generalizable Multi-source Person Re-identification}

\author{Lei Qi,
        Lei Wang,
        Yinghuan Shi,
        Xin Geng$^*$
\thanks{This work was supported in part by the National Key Research and Development Plan of China under Grant (2018AAA0100104), in part by Jiangsu Natural Science Foundation Project (BK20210224), the National Science Foundation of China under Grant (62125602), Grant (62076063), NSFC Major Program (62192783), CAAI-Huawei MindSpore Project (CAAIXSJLJJ-2021-042A) and China Postdoctoral Science Foundation Project (2021M690609).}

\thanks{Lei Qi and Xin Geng are with the School of Computer Science and Engineering, and the Key Lab of Computer Network and Information Integration (Ministry of Education), Southeast University, Nanjing, China, 211189 (e-mail: qilei@seu.edu.cn; xgeng@seu.edu.cn).}
\thanks{Lei Wang is School of Computing and Information Technology, University of Wollongong, Wollongong, Australia (e-mail: leiw@uow.edu.au).}
\thanks{Yinghuan Shi is with the State Key Laboratory for Novel Software Technology, Nanjing University, Nanjing, China, 210023 (e-mail: syh@nju.edu.cn).}
\thanks{* Corresponding author: Xin Geng.}
}

%
%

\markboth{ }%
{Shell \MakeLowercase{\textit{et al.}}: Bare Demo of IEEEtran.cls for IEEE Journals}

\maketitle

\begin{abstract}
Person re-identification (Re-ID) has achieved great success in the supervised scenario. However, it is difficult to directly transfer the supervised model to arbitrary unseen domains due to the model overfitting to the seen source domains. In this paper, we aim to tackle the generalizable multi-source person Re-ID task (\ie, there are multiple available source domains, and the testing domain is unseen during training) from the data augmentation perspective, thus we put forward a novel method, termed MixNorm. It consists of domain-aware mix-normalization (DMN) and domain-aware center regularization (DCR). Different from the conventional data augmentation, the proposed domain-aware mix-normalization enhances the diversity of features during training from the normalization perspective of the neural network, which can effectively alleviate the model overfitting to the source domains, so as to boost the generalization capability of the model in the unseen domain. To further promote the efficacy of the proposed DMN, we exploit the domain-aware center regularization to better map the diversely generated features into the same space. Extensive experiments on multiple benchmark datasets validate the effectiveness of the proposed method and show that the proposed method can outperform the state-of-the-art methods. Besides, further analysis also reveals the superiority of the proposed method.
\end{abstract}


%
\IEEEpeerreviewmaketitle

\section{Introduction}
\IEEEPARstart{I}{n} recent years, person re-identification (Re-ID) has attracted an increasing interest in both academia and industry due to its great potentials in the video surveillance system~\cite{zheng2016person,ye2020deep,DBLP:journals/tcsv/LengYT20}. Person Re-ID resorts to matching images of the same person captured by different cameras with the non-overlapping camera views. This main challenge of person Re-ID is the variations including body pose, viewing angle, illumination, image resolution, occlusion, background and so on across different cameras~\cite{DBLP:journals/pami/LiZG20,chen2021occlude,DBLP:journals/tcsv/QiWHSG20}. In general, person Re-ID can be treated as a special case of the image retrieval problem with the goal of querying from a large-scale gallery set to quickly and accurately find images that match with a query image.

Currently, the typical person Re-ID methods have obtained good performance in the supervised setting due to the power of deep neural network~\cite{DBLP:journals/tmm/ZhaoLZZWM20,DBLP:journals/tmm/WeiZY0019,DBLP:conf/cvpr/WuZGL19,DBLP:conf/cvpr/LiZG18,DBLP:conf/cvpr/ZhengYY00K19,DBLP:journals/tomccap/QiWHSG21,hou2020iaunet,hou2021feature}. For example, in \cite{hou2020iaunet}, a novel 
interaction–aggregation-update is introduced to comprehensively leverage the spatial–temporal
context information, which can enable the feature to incorporate the globally spatial, temporal, and channel context. To address the occluded person Re-ID task, Hou \etal~\cite{hou2021feature} design an occlusion-robust block, which can effectively recover the semantics of occluded regions in the feature space. 
However, when these models are used to the unseen domain, they will fail on performance because of the model overfitting to the source domains. Generally, when using these methods in a new domain or scenario, we need to collect the data and then label the data, which is expensive and time-consuming, thus these existing supervised models cannot be directly deployed in the real-world application. Although some unsupervised domains adaptation (UDA) methods are proposed to mitigate the labeling task~\cite{DBLP:journals/tmm/YangYLJXYGHG21,DBLP:conf/cvpr/ZhaiLYSCJ020,DBLP:conf/iccv/WuZL19,DBLP:conf/eccv/ChenLL020,DBLP:conf/iccv/QiWHZSG19}, they still require to collect data and re-train the model for the new scenario. 

Domain generalization (DG) methods can address the above problem, which resorts to learning a model in the source domains and testing in the unseen domain~\cite{zhou2021domain}. In the person Re-ID community, some DG methods have been developed to produce a robust model for the unseen target domain. For example, the method in~\cite{DBLP:conf/eccv/LiaoS20} treats image matching as finding local correspondences in feature maps and constructs query-adaptive convolution kernels on the fly to achieve local matching. In~\cite{DBLP:conf/cvpr/ZhaoZYLLLS21}, a meta-learning strategy is introduced to simulate the train-test process of domain generalization for learning a more generalizable model. The method \cite{DBLP:conf/cvpr/DaiLLTD21} adopts an effective voting-based mixture mechanism to dynamically leverage the diverse characteristics of source domains to improve the generalization ability of the model. Differently, we aim to solve the issue from the data-augmentation perspective.

In this paper, we focus on the generalizable multi-source person re-identification task as in~\cite{DBLP:conf/cvpr/ZhaoZYLLLS21}, where there are multiple available source domains in the training stage, and the testing data is unseen during training. Considering the conventional batch normalization (BN)~\cite{DBLP:conf/icml/IoffeS15} uses the same statistics to normalize all samples in a batch, which could lead to learning the fixed pattern in the training process and further overfitting the trained model to the seen source domains. To address the issue, we propose a domain-aware mix-normalization (DMN) to achieve the feature augmentation, which can alleviate the overfitting problem via yielding the diverse features. To be specific, we randomly combine different domains to generate the mixed statistics (\ie, mean and variance) in a batch when conducting the forward operation during training, and they are employed to normalize the samples in the combined domains. To deeply mine the power of the proposed DMN, we employ the domain-aware center regularization (DCR)~\cite{DBLP:conf/cvpr/GuoZZCLL20} to better train the model with the diverse features, which can effectively map all features into the same space, so that the domain-invariant feature can be obtained. Together, the above two modules give rise to a novel ``MixNorm'' method for generalizable multi-source person Re-ID. We conduct the experiments on multiple benchmark person Re-ID datasets to confirm the effectiveness of the proposed method. Besides, the deep analysis by extensive experiments reveals the superiority of MixNorm.
In this paper, our main contributions can be summarized as:
  \begin{itemize}
    \item We propose a simple yet effective domain-aware mix-normalization, which can produce diverse features to enhance the robustness and effectiveness of the model in unseen domains. 
    \item 
    To further promote the power of our proposed mix-normalization during training, we exploit a domain-aware center regularization to guarantee that the diverse samples from DMN are mapped into the same space.
    \item We evaluate our approach on multiple standard benchmark datasets, and the results show that our approach outperforms the state-of-the-art accuracy. Moreover, the ablation study and further analysis are provided to validate the efficacy of our method.
  \end{itemize}

The rest of this paper is organized as follows.
We review some related work in Section \ref{s-related}.
The proposed method is introduced in Section \ref{s-framework}.
Experimental results and analysis are presented in Section \ref{s-experiment},
and Section \ref{s-conclusion} is conclusion.

\section{Related work}\label{s-related}
In this section, we review the related work to our work, including the generalizable person re-identification, the domain generalization methods and open set domain adaptation. The detailed investigation is presented in the following part.

\subsection{Generalizable Person Re-identification}
Currently, several methods have been developed to solve the generalizable (\ie, domain generalization) person Re-ID task, including the instance normalization based methods, the meta-learning based methods and the sample-adaptive methods.

Considering that the domain discrepancy in Re-ID is due to style and content variance across datasets, and instance normalization (IN)~\cite{ulyanov2016instance} can effectively remove the style information,  Jia \etal \cite{DBLP:conf/bmvc/JiaRH19} adopt instance and feature normalization to alleviate much of the resulting domain-shift in deep Re-ID models. However, IN inevitably removes discriminative information, while it filters out style variations, such as illumination and color contrast. Therefore, Jin \etal~\cite{DBLP:conf/cvpr/JinLZ0Z20}  propose
to distill identity-relevant feature from the removed information and restitute it to the network to ensure high discrimination. For better disentanglement, the method enforces a dual
causality loss constraint in SNR to encourage the separation of identity-relevant features and identity-irrelevant features.

Besides, some methods employ the meta-learning scheme to enhance the generalization capability to the unseen domains. In \cite{DBLP:conf/cvpr/SongYSXH19}, a novel deep Re-ID model termed Domain-Invariant
Mapping Network (DIMN) is proposed, which utilizes the meta-learning pipeline to sample a subset of source domain training tasks during each training episode, so as to make the model domain-invariant.
Choi \etal~\cite{DBLP:conf/cvpr/ChoiKJPK21} 
combine learnable batch-instance normalization layers with meta-learning and investigate the challenging cases caused by both batch and instance normalization layers.  Moreover, the method diversifies the virtual simulations via the meta-train loss accompanied by a cyclic
inner-updating manner to boost the model's generalization.
Zhao \etal~\cite{DBLP:conf/cvpr/ZhaoZYLLLS21} design the Memory-based Multi-Source MetaLearning (M$^3$L) framework to train a generalizable model
for unseen domains. In this framework, a meta-learning strategy is introduced to simulate the train-test process of domain generalization for learning more generalizable models. Moreover, this method also presents a meta batch normalization layer to diversify meta-test features, further establishing the advantage of meta-learning.

Different from the aforementioned methods, some adaptive approaches are also designed to perform well in the unseen target domain, which can adjust the parameters or module according to each testing image in the testing stage. For example, Dai \etal~\cite{DBLP:conf/cvpr/DaiLLTD21} introduce a novel method called the relevance-aware mixture of experts (RaMoE), using an effective voting-based mixture mechanism to dynamically leverage the diverse characteristics of source domains to improve the generalization ability of the model. Particularly, in the testing stage, the method can change the weights of each experts according to the current testing image.
In~\cite{DBLP:conf/eccv/LiaoS20},  the method treats image matching as finding local correspondences in feature maps, and constructs query-adaptive convolution kernels on the fly to
achieve local matching. In this way, the matching process and results are interpretable, and this explicit matching is more generalizable than representation features to unseen domains, such as unknown misalignments,
pose or viewpoint changes. 

In this paper, unlike the above methods for generalizable person Re-ID, we resort to improving the generalization capability of the model from the data augmentation perspective when there are multiple available source domains during training, as the setting in~\cite{DBLP:conf/cvpr/ZhaoZYLLLS21}.

\subsection{Domain Generalization}
Recently, some methods are also proposed to address the domain generalization problem in the classification and semantic segmentation tasks~\cite{DBLP:conf/cvpr/NamLPYY21,DBLP:conf/eccv/SeoSKKHH20,DBLP:conf/iccv/YueZZSKG19,DBLP:conf/cvpr/CarlucciDBCT19,DBLP:conf/nips/BalajiSC18,DBLP:conf/iccv/LiZYLSH19,DBLP:conf/eccv/LiTGLLZT18,DBLP:journals/pr/ZhangQSG22}. Particularly, the data augmentation based methods have obtained significant advance in these tasks, which can be divided into image-level and feature-level data augmentation.


For the image-level augmentation, Zhou \etal \cite{DBLP:conf/eccv/ZhouYHX20} employ a data generator to synthesize data from pseudo-novel domains to augment the source domains. Moreover,
in \cite{DBLP:conf/aaai/ZhouYHX20},  a
novel DG approach based on Deep Domain-Adversarial Image Generation (DDAIG) is proposed, which aims to map the source training data into unseen domains. This is achieved by a learning objective formulated to ensure that the generated data can be correctly classified by the label classifier while fooling the domain classifier. Particularly, augmenting the source training data with the generated
unseen domain data can make the label classifier more robust to unknown domain changes. Differently, considering the Fourier phase information contains high-level semantics and is not easily
affected by domain shifts,
Xu~\etal \cite{DBLP:conf/cvpr/XuZ0W021} develop a novel Fourier-based data
augmentation strategy to force the model to capture
phase information, which linearly
interpolates between the amplitude spectrums of two images.

For the feature-level augmentation, 
Huang \etal \cite{DBLP:conf/eccv/HuangWXH20} 
introduce a simple training heuristic, called Representation Self-Challenging
(RSC), that significantly improves the generalization of CNN to the out-of-domain data. RSC iteratively discards the dominant features activated on the training data, and forces the network to activate
remaining features that correlate with labels.
Li \etal~\cite{Li_2021_ICCV} propose an extremely simple technique of perturbing
the feature embedding with Gaussian noise during training leads to a classifier with domain-generalization performance comparable to existing state-of-the-art. 
Besides, in \cite{DBLP:conf/iclr/ZhouY0X21}, a novel approach is proposed based on probabilistically mixing instance-level feature statistics of training samples across source domains,
termed MixStyle.  Mixing styles of training instances results in novel domains being synthesized implicitly, which increases the domain diversity of the source domains, and
hence improves the generalization ability of the trained model.

In this paper, we solve the generalizable person Re-ID task from the data augmentation perspective. Particularly, our proposed method belongs to the feature-level augmentation. Differently, we achieve to generate the diverse features from the normalization view of the neural network, which is not well investigated in the existing works.

\subsection{Open Set Domain Adaptation}
Generalizable person re-identification can be also considered as an open set task, where there is no overlapping category between source domains and unseen target domain. Recently, the open set setting has attracted much attention in the domain adaptation community, which is firstly proposed in \cite{DBLP:conf/iccv/BustoG17}. Some methods have been designed to deal with the issue~\cite{DBLP:conf/eccv/SaitoYUH18,DBLP:conf/cvpr/LiuCLW019,DBLP:conf/cvpr/PanYLNM20,DBLP:conf/icml/LuoWHB20,DBLP:journals/tmm/SherminLTMS21}. For example,
Saito~\etal~\cite{DBLP:conf/eccv/SaitoYUH18} propose to utilize adversarial training to extract features that separate unknown target from known target samples. 
In~\cite{DBLP:conf/cvpr/LiuCLW019}, the approach adopts a coarse-to-fine weighting mechanism to
progressively separate the samples of unknown and known classes, and simultaneously measure their importance on feature distribution alignment. 
Luo~\etal~\cite{DBLP:conf/icml/LuoWHB20} introduce an end-to-end progressive graph learning framework where a graph neural network with episodic training is integrated to suppress underlying conditional shift, and adversarial learning is adopted to close the gap between the source and target distributions.
Moreover, Shermin~\etal~\cite{DBLP:journals/tmm/SherminLTMS21} put forward a novel adversarial domain adaptation model with multiple auxiliary classifiers, which can effectively encourage positive transfers during adversarial training and simultaneously reduce the domain gap between the shared classes of the source and target domains.

Different from the open set domain adaptation task, the target domain is unseen during the training course in the generalizable person Re-ID task. Therefore, we cannot directly pull the distance between source and target domains in the generalizable Re-ID task. In this paper, we solve this issue by generating the diverse features to enrich the data diversity from the normalization perspective.


\begin{figure*}[t]
\centering
\includegraphics[width=16cm]{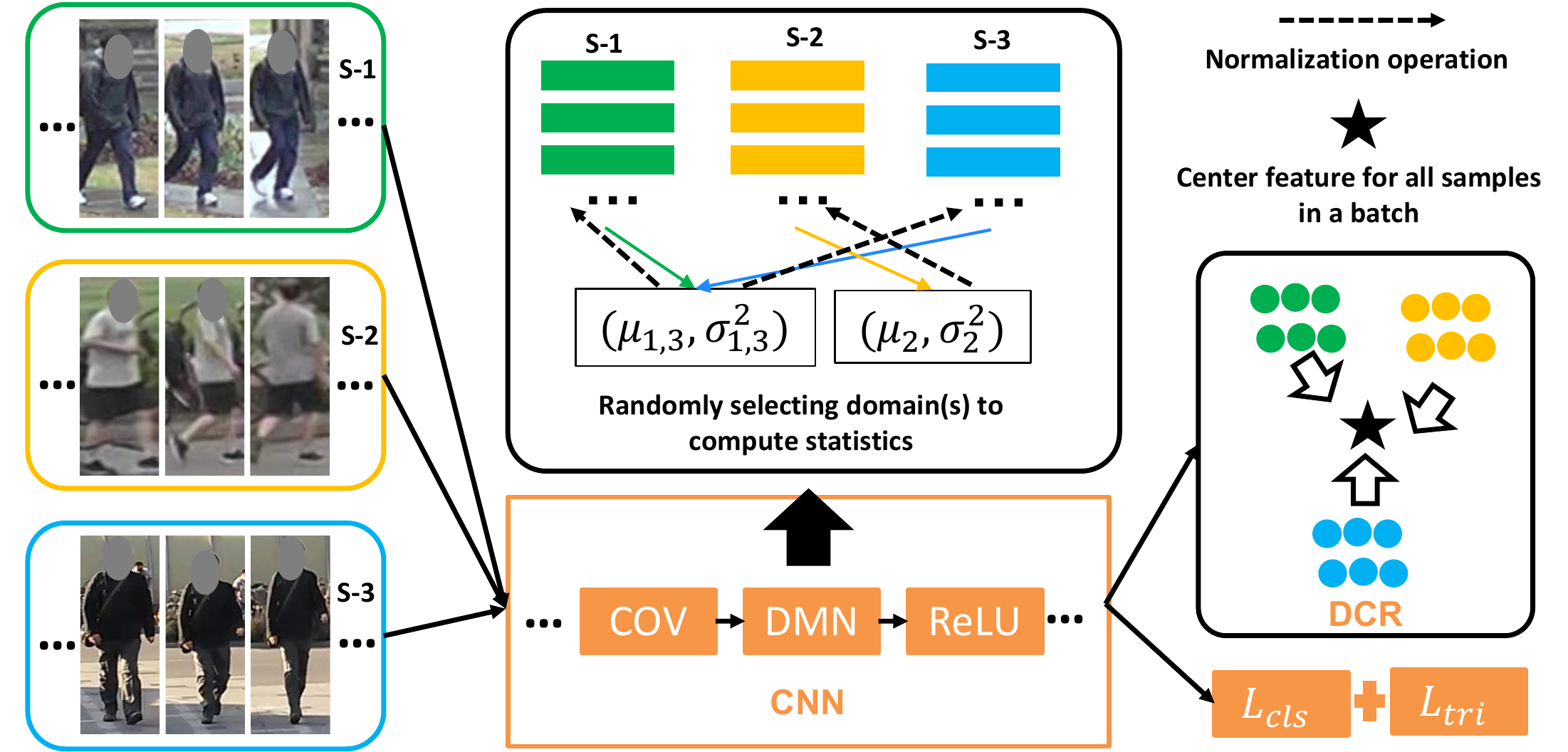}
\caption{An illustration of the proposed MixNorm. Here we take three source domains as an example. As seen in this figure, the method is composed of domain-aware mix-normalization (DMN) and domain-aware center regularization (DCR). In this figure, we show that source-1 and source-3 are randomly selected to calculate statistics for normalization together, and the remaining source-2 independently conducts the normalization operation. These green, yellow and blue blocks in ``S-1'', ``S-2'' and ``S-3'' represent the feature map for each sample of the corresponding domain, which will be normalized by our DMN. Best viewed in color.}
\label{fig04}
\end{figure*}

\section{The proposed method}\label{s-framework}

In this paper, to enhance the generalization capacity of the model to the unseen domain, we put forward a novel mix-normalization method termed MixNorm, as illustrated in Fig.~\ref{fig04}. Specifically, in this method, we develop domain-aware mix-normalization (DMN) to conduct the feature augmentation, which randomly combines different domains to compute the statistics during each normalization. Thus, perturbed features can be generated to promote the diversity of the training samples. Furthermore, to boost the effectiveness of the DMN, we leverage the domain-aware center regularization (DCR)~\cite{DBLP:conf/cvpr/GuoZZCLL20} to map all samples into the same space, which can effectively enforce the model to learn the domain-invariant features. The detailed description of the proposed method will be presented in the following part.

\subsection{Domain-aware Mix-normalization}
In this part, we will introduce the domain-aware mix-normalization (DMN) for generalizable multi-source person re-identification. During training, we randomly select the same number of samples from each source domain to form each batch. For example, we have $D$ source domains and choose $P$ samples from each domain, thus the batch size is $N = D\times P$ in the training stage. In general, the conventional batch normalization (BN)~\cite{DBLP:conf/icml/IoffeS15} is leveraged to normalize all samples in each batch, which can be defined as:
\begin{equation}
  \begin{aligned}
  &{\rm BN}(f_d)= \gamma \frac{f_d-\mu}{\sigma}+\beta,~~d \in \{1, ..., D\},
  \end{aligned}
  \label{eq01}
  \end{equation}
  where $f_d \in \mathbb{R}^{N\times C\times H \times W}$ denotes the feature maps from the $d$-th domain, $\gamma,\beta \in \mathbb{R}^{C}$ are learnable affine transformation parameters, and $C$ is the number of channel on the corresponding feature maps. $\mu, \sigma \in \mathbb{R}^{C}$ represent the channel-wise mean and standard deviation of BN for the feature maps (\ie, statistics) as follows:
  \begin{equation}
  \mu=\frac{1}{DPHW}\sum_{d=1}^{D}\sum_{p=1}^{P}\sum_{h=1}^{H}\sum_{w=1}^{W}f_d[p,:, h,w],
    \label{eq02}
  \end{equation}
  \begin{equation}
  \sigma=\sqrt{\frac{1}{DPHW}\sum_{d=1}^{D}\sum_{p=1}^{P}\sum_{h=1}^{H}\sum_{w=1}^{W}(f_d[p,:,h,w]-\mu)^2 + \epsilon},
  \label{eq03}
  \end{equation}
  where $\epsilon$ is a constant for numerical stability. As seen in the above equation, BN utilizes the same statistics (\ie,  a pair of $\mu$ and $\sigma$) for all samples in each batch, which could result in that the fixed pattern is learned by neural network, so that the final model overfits to the source domains. Most existing methods address the issue in person Re-ID community by the instance normalization based methods, the meta-learning based methods or the sample-adaptive methods. 

  Differently, we propose to tackle this problem from the data augmentation perspective based on the normalization of the neural network. Considering that using the same mean and variance for all samples in each batch could capture the fixed pattern, we propose a domain-aware mix-normalization for performing the normalization, which randomly combines the different domains to yield the diverse statistics. Therefore, normalizing the features using diverse statistics can achieve the function of the feature-level data augmentation. The domain-aware mix-normalization (DMN) can be formulated by:
    \begin{equation}
  \begin{aligned}
  &{\rm DMN}(f_d)= \gamma \frac{f_d-M_\mu(\phi)}{M_{\sigma}(\phi)}+\beta,~~d \in \phi,
  \end{aligned}
    \label{eq04}
  \end{equation}
  where $\phi$ is the randomly selected domain set, and $M_{\mu}(\phi)$ and $M_{\sigma}(\phi)$ represent the statistics for the $\phi$. For the DMN,  we firstly initialize a domain set $\mathbf{S}$ including all domains and produce a random integer $C$, which is constrained between $1$ and $D-1$ (\ie, $[1,  D-1]$) in our method. 
  Then we randomly choose $C$ domains (\ie, $\phi$) from $\mathbf{S}$ and remove them from $\mathbf{S}$. For the selected domain set $\phi$, we compute the statistics as Eqs.~\ref{eq05} and~\ref{eq06} and conduct the normalization as Eq.~\ref{eq04}. We will continue to select $C$ domains from $\mathbf{S}$ until $\mathbf{S}$ is empty. It is worth noting that, if the number of the remaining domains is smaller than $C$, we will leverage all remaining domains to compute the statistics for normalizing themselves. Specifically, $M_{\mu}(\phi)$ and $M_{\sigma}(\phi)$  can be described as:
  


\begin{equation}
  M_{\mu}(\phi)=\frac{1}{|\phi|PHW}\sum_{d\in \phi}\sum_{p=1}^{P}\sum_{h=1}^{H}\sum_{w=1}^{W}f_{d}[p,:, h,w],
  \label{eq05}
  \end{equation}
  \begin{equation}
  \begin{aligned}
     & M_{\sigma}(\phi)= \\
      &\sqrt{\frac{1}{|\phi|PHW}\sum_{d \in \phi}\!\sum_{p=1}^{P}\!\sum_{h=1}^{H}\!\sum_{w=1}^{W}(f_{d}[p,:,h,w]\!-\!M_{\mu}(\phi))^2\!+\!\epsilon},
  \end{aligned}
  \label{eq06}
  \end{equation}
  where $|\phi|$ denotes the number of the selected domains.
  The forward process of the proposed DMN is given in Algorithm~\ref{al01}. Moreover,
  it is worth noting that there is only a pair of statistics for evaluation in each DMN during testing, \ie, the statistics of all random combinations are accumulated to a pair of the mean and variance used for the evaluation, as the operation in BN~\cite{DBLP:conf/icml/IoffeS15}.

\textit{Remark.} For the proposed DMN, we use it by replacing all the original BN layers in the whole network. Particularly, since each DMN is independent, the randomization appears not only in each normalization but also between normalizations of different layers. This way can bring more diversity of features during training so as to promote the generalization to the unseen domain, as validated in the further analysis of the experimental section.

Besides, when using the DMN to train the model, we set the maximum numbers of the randomly selected domains as $D-1$. If it is set as $D$ (\ie, using the mean and variance of all domains to normalize features during training), it could cause slight overfitting to the source domains, because the final statistics for testing are the expectation of all domains. In other words, if all domains are selected together, this will be equivalent to the raw batch normalization, which cannot diversify features in the normalization layer. Therefore, this could result in the overfitting to source domains. The related results are given in the experiment.

Moreover, considering that the DMN requires the uniform sampling for each domain during the training course, \ie, there is the same number of samples for each domain in a batch, which is short of the randomization when compared with the random sampling, \ie, we combine all samples from all domains to a dataset and then randomly sample from the combined dataset to generate each batch. Particularly, although the random sampling scheme can bring the randomization across different batches, it still utilizes the global statistics to normalize all samples in each batch. Differently, our method randomly produces the local statistics in each batch to normalize the different samples, thus it could better prevent the model from learning the fixed pattern, so as to alleviate the overfitting to the seen source domains. In the experimental section, we also verify that our method can produce a more robust model than the random sampling scheme in the generalizable multi-source person Re-ID task.

Particularly, our DMN is different from MetaBN~\cite{DBLP:conf/cvpr/ZhaoZYLLLS21}. The original MetaBN mixes the original meta-test features and the sampled features to diversify the the original meta-test features, and then performs the batch normalization for the mixed features. Differently, our method randomly mixes the statistics of different domains, and then uses the statistics for normalizing the original features. The advantage of our method is that, all samples from different domains in each batch are not normalized by the shared statistics as the conventional batch normalization, thus it is beneficial for yielding diverse features.



\begin{algorithm}[ht]
\caption{\small{The forward process of mix-normalization (DMN)}}~\label{alg01}
\begin{algorithmic}[1]
\STATE {\bf Input:} 
Feature $f$ of all $D\times P$ samples in a batch.\\
\STATE {\bf Output:} The normalized feature $\hat{f}$. \\
\STATE {\bf Initialization:} The domain set $\mathbf{S}$ is all $D$ source domains. \\
\STATE Randomly produce the number of mixing domains as $C$.
\WHILE{$|\mathbf{S}|\neq 0$}
\IF {$|\mathbf{S}|=<C$}
\STATE Set $C$ as $|\mathbf{S}|$.
\ENDIF
\STATE Randomly select $C$ domains $\phi$ from the $\mathbf{S}$. \\
\STATE Remove the selected $C$ domains  (\ie, $\phi$) from the $\mathbf{S}$.\\
\STATE Compute the statistics (\ie, mean and variance) for the selected $C$ domains as Eqs.~\ref{eq05} and ~\ref{eq06}.\\
\STATE Utilize the statistics to normalize all samples from the selected $C$ domains as Eq.~\ref{eq04}.
\ENDWHILE
\end{algorithmic}
\label{al01}
\end{algorithm}

\subsection{Domain-aware Center Regularization}
The proposed domain-aware mix-normalization can enhance the diversity of features during training, thus we further utilize domain-aware center regularization (DCR)~\cite{DBLP:conf/cvpr/GuoZZCLL20} to better learn the domain-invariant model using the diverse features. Generally, the Maximum Mean Discrepancy (MMD)~\cite{DBLP:journals/jmlr/GrettonBRSS12} can be used to reduce domain shift between two domains. In our task, there are available multiple domains in the training process, thus the MMD cannot be directly employed to address our issue. Therefore, we utilize the domain-aware center regularization~\cite{DBLP:conf/cvpr/GuoZZCLL20} to reduce the domain gap across different domains, which can be defined as:
  \begin{equation}
  \begin{aligned}
  L_{dcr}=\sum_{d=1}^D\sum_{p=1}^P\|r[d\times p, :]-\bar{r}\|_2^2,
  \end{aligned}
  \label{eq07}
  \end{equation}
  where $r \in \mathbb{R}^{N \times B}$ is the feature representation of all samples in a batch, and $B$ indicates the dimension of the the features. $\bar{r}\in \mathbb{R}^{1 \times B}$ denotes the mean of all features.

  \textit{Remark.} It is worth noting that the domain-aware center regularization is merely effective when the domain-aware mix-normalization is used. The main reason is that the original model is easy to train due to the single pattern (\ie, using the same mean and variance for all samples in a batch) in the training stage. Since using the domain-aware mix-normalization introduces the diverse features, it could need a strong regularization to map all features into the same space, so as to learn the domain-invariant model. We will report the results in the experimental section.

\subsection{The Overall Loss}
During training, we employ the cross-entropy loss (\ie, $L_{cls}$), the triplet loss (\ie, $L_{tri}$) with hard mining sampling~\cite{ulyanov2016instance} and the domain-aware center regularization (\ie, $L_{dcr}$)  to train the model. Particularly, the cross-entropy loss and the triplet loss are the basic loss in the person Re-ID community~\cite{DBLP:journals/tmm/LuoJGLLLG20,DBLP:conf/iccv/FuWWZSUH19}. The overall loss for training the model can be described as:
  \begin{equation}
  \begin{aligned}
  L_{overall}=L_{cls}+L_{tri}+\lambda L_{dcr},
  \end{aligned}
  \label{eq08}
  \end{equation}
  where $\lambda$ is the hyper-parameter to trade off the basic loss and the domain-aware center regularization.
\section{Experiments}\label{s-experiment}
\renewcommand{\cmidrulesep}{0mm} 
\setlength{\aboverulesep}{0mm} 
\setlength{\belowrulesep}{0mm} 
\setlength{\abovetopsep}{0cm}  
\setlength{\belowbottomsep}{0cm}

In this part, we firstly introduce the experimental datasets and settings in Section~\ref{sec:EXP-DS}. Then, we compare the proposed method with the state-of-the-art generalizable Re-ID methods in Section~\ref{sec:EXP-CUA}, respectively. To validate the effectiveness of various components in the proposed framework, we conduct ablation studies in Section~\ref{sec:EXP-SS}. Lastly, we further analyze the property of the proposed method in Section~\ref{sec:EXP-FA}.
\subsection{Datasets and Experimental Settings}\label{sec:EXP-DS}
\subsubsection{Datasets} 
We evaluate our approach on four large-scale image datasets: Market1501~\cite{DBLP:conf/iccv/ZhengSTWWT15}, DukeMTMC-reID~\cite{DBLP:conf/iccv/ZhengZY17}, MSMT17~\cite{wei2018person} and CUHK03-NP~\cite{DBLP:conf/cvpr/LiZXW14,DBLP:conf/cvpr/ZhongZCL17}. 
 \textbf{Market1501 (Ma)} contains 1,501 persons with 32,668 images from six cameras. Among them, $12,936$ images of $751$ identities are used as a training set. For evaluation, there are $3,368$ and $19,732$ images in the query set and the gallery set, respectively. \textbf{DukeMTMC-reID (D)} has $1,404$ persons from eight cameras, with $16,522$ training images, $2,228$ query images, and $17,661$ gallery images.
 \textbf{MSMT17 (Ms)} 
 is collected from a 15-camera network deployed on campus. The training set contains $32,621$ images of $1,041$ identities. For evaluation, $11,659$ and $82,161$ images are used as query and gallery images, respectively. \textbf{CUHK03-NP (C)} has an average of 4.8 images in each camera. The dataset provides both manually labeled bounding boxes and DPM-detected bounding boxes. On this dataset, there are $7,365$ training images, and $1,400$ images and $5,332$ images in query set and gallery set are used in the testing stage. Particularly, we divide these four datasets into two parts: three domains as source domains for training and the other one as target domain for testing. We adopt the recommended setting in ~\cite{DBLP:conf/cvpr/ZhaoZYLLLS21}.

In addition, we also train our method on more datasets including Market1501, DukeMTMC-reID, CUHK03, CUHK02~\cite{DBLP:conf/cvpr/LiW13} and CUHK-SYSU~\cite{xiao2016end}, and test on two small-scale datasets including PRID~\cite{DBLP:conf/scia/HirzerBRB11} and VIPeR~\cite{DBLP:conf/eccv/GrayT08}. Particularly, the performances of these small ReID datasets are evaluated on the average of 10 repeated random splits of gallery and probe sets. 
   For all datasets, we employ CMC (\ie, Cumulative Match Characteristic) accuracy and mAP (\ie, mean Average Precision) for Re-ID evaluation~\cite{DBLP:conf/iccv/ZhengSTWWT15}.

\subsubsection{Implementation Details} 
In this experiment, we use the ResNet-50~\cite{DBLP:conf/cvpr/HeZRS16} and IBN-Net50~\cite{DBLP:conf/eccv/PanLST18} pre-trained on ImageNet~\cite{DBLP:conf/cvpr/DengDSLL009} to initialize the network parameters. In a batch, the number of IDs and the number of images per person are set as $16$ and $4$ to produce triplets for each domain, respectively.
 The initial learning rate is $3.5\times 10^{-4}$ and divided by $10$ at the $30$-th and $50$-th epochs, respectively. The proposed model is trained with the Adam optimizer in a total of $60$ epochs. The size of the input image is $256 \times 128$. For data augmentation, we perform random cropping,
random flipping and auto-argumentation~\cite{cubuk2018autoaugment}. Besides, $\lambda$ in Eq.~\ref{eq08} is set as $0.2$. 
 Particularly, we utilize the same setting for all experiments on all datasets in this paper.

\subsection{Comparison with State-of-the-art Methods}\label{sec:EXP-CUA}
We compare our proposed method (\ie, MixNorm) with some state-of-the-art methods as reported in Table~\ref{tab02}, including IBN-Net50~\cite{DBLP:conf/eccv/PanLST18}, QAConv$_{50}$~\cite{DBLP:conf/eccv/LiaoS20}, and M$^3$L~\cite{DBLP:conf/cvpr/ZhaoZYLLLS21}. IBN-Net50 carefully integrates Instance Normalization (IN) and Batch Normalization (BN) as building blocks, where IN learns
features that are invariant to appearance changes, such as colors and styles, while BN is essential for preserving content-related
information. This method remarkably enhances a CNN’s modeling ability on one domain as well as its generalization capacity on another domain without finetuning.
QAConv$_{50}$~treats image matching as finding local correspondences in feature maps, and constructs query-adaptive convolution kernels on the fly to achieve local matching.
M$^3$L utilizes a meta-learning strategy to simulate the train-test process of domain generalization for learning more generalizable models. 
As seen in Table~\ref{tab02}, our MixNorm outperforms all other methods on Rank-1 and mAP. For example, in the ``D+C+Ms$\rightarrow$Ma'' task, our method increases M$^3$L by $+2.2\%$ ($54.7$ vs. $52.5$) and $+3.1\%$ ($81.4$ vs. $78.3$) on mAP and Rank-1 when using IBN-Net50 as the backbone. This mainly owes to the effectiveness of the domain-aware mix-normalization, which can bring diverse features to prevent the model's overfitting to the source domains. Moreover, using domain-aware center regularization can better exploit these diverse features to learn the domain-invariant model. Besides, although the performance of our method is slightly poorer than M$^3$L when ResNet-50 is used as the backbone in the ``Ma+D+Ms$\rightarrow$C'' task, our method has better results than M$^3$L based on IBN-Net50. Particularly, in this task, each component of our method is also effective, as shown in the below ablation study.

\begin{table}[htbp]
  \centering
  \caption{Experimental results on four large-scale datasets, including Market1501 (Ma), DukeMTMC-reID (D), MSMT17 (Ms), and CUHK03 (C). ``D+C+Ms$\rightarrow$Ma'' indicates that the model is trained on D, C and Ms, and tested on Ma. The bold denotes the best result.}
    \begin{tabular}{l|cc|cc}
    \toprule
\multicolumn{1}{c|}{\multirow{2}[1]{*}{Method}} & \multicolumn{2}{c|}{D+C+Ms$\rightarrow$Ma} & \multicolumn{2}{c}{Ma+C+Ms$\rightarrow$D} \\
\cmidrule{2-5}  & mAP   & Rank-1 & mAP   & Rank-1 \\
    \midrule
    IBN-Net50~\cite{DBLP:conf/eccv/PanLST18} & 43.0  & 73.4  & 45.7  & 64.9  \\
    QAConv$_{50}$~\cite{DBLP:conf/eccv/LiaoS20} & 39.5  & 68.6  & 43.4  & 64.9  \\
    M$^3$L\scriptsize{(ResNet-50)}~\cite{DBLP:conf/cvpr/ZhaoZYLLLS21} & 51.1  & 76.5  & 48.2  & 67.1  \\
    M$^3$L\scriptsize{(IBN-Net50)}~\cite{DBLP:conf/cvpr/ZhaoZYLLLS21} & 52.5  & 78.3  & 48.8  & 67.2  \\
    \midrule
    MixNorm \scriptsize{(ResNet-50)} & 51.4  & 78.9  & 49.9  & \textbf{70.8} \\
    MixNorm \scriptsize{(IBN-Net50)} & \textbf{54.7} & \textbf{81.4} & \textbf{52.3} & 70.6  \\
    \midrule
    \multicolumn{1}{c|}{\multirow{2}[1]{*}{Method}} & \multicolumn{2}{c|}{Ma+D+C$\rightarrow$Ms} & \multicolumn{2}{c}{Ma+D+Ms$\rightarrow$C} \\
\cmidrule{2-5}          & mAP   & Rank-1 & mAP   & Rank-1 \\
    \midrule
    IBN-Net50~\cite{DBLP:conf/eccv/PanLST18} & 17.0  & 43.9  & 21.3  & 21.6  \\
    QAConv$_{50}$~\cite{DBLP:conf/eccv/LiaoS20} & 10.0  & 29.9  & 19.2  & 22.9  \\
    M$^3$L\scriptsize{(ResNet-50)}~\cite{DBLP:conf/cvpr/ZhaoZYLLLS21} & 13.1  & 32.0  & 30.9  & 31.9  \\
    M$^3$L\scriptsize{(IBN-Net50)}~\cite{DBLP:conf/cvpr/ZhaoZYLLLS21} & 15.4  & 37.1  & 31.4  & 31.6  \\
    \midrule
    MixNorm \scriptsize{(ResNet-50)} & 19.4  & 47.2  & 29.0  & 29.6  \\
    MixNorm \scriptsize{(IBN-Net50)} & \textbf{23.1} & \textbf{52.2} & \textbf{32.3} & \textbf{32.4} \\
    \bottomrule
    \end{tabular}%
  \label{tab02}%
\end{table}%

In addition, our method is also compared with some other methods on small-scale datasets when using more datasets to train the model, and these methods include
Agg Align~\cite{zhang2017alignedreid},
    Reptile~\cite{nichol2018first},
    CrossGrad~\cite{DBLP:conf/iclr/ShankarPCCJS18}, 
    Agg PCB~\cite{DBLP:journals/pami/SunZLYTW21},
    MLDG~\cite{DBLP:conf/aaai/LiYSH18},
    PPA~\cite{DBLP:conf/cvpr/QiaoLSY18},
    DIMN~\cite{DBLP:conf/cvpr/SongYSXH19},
    SNR~\cite{DBLP:conf/cvpr/JinLZ0Z20} and RaMoE~\cite{DBLP:conf/cvpr/DaiLLTD21}.
The experimental results are given in Table~\ref{tab01}. For example, DIMN~\cite{DBLP:conf/cvpr/SongYSXH19}  learns a mapping between a person image and its identity classifier and follows a
meta-learning pipeline to sample a subset of source domain training tasks during each training episode.
    SNR~\cite{DBLP:conf/cvpr/JinLZ0Z20} distills identity-relevant features from the removed information conducted by instance normalization and restitute it to the network to ensure high discrimination. RaMoE \cite{DBLP:conf/cvpr/DaiLLTD21} adopts an effective voting-based mixture mechanism to dynamically leverage the diverse characteristics of source domains to improve the model’s generalization. 
  As seen in Table~\ref{tab01}, our method can basically obtain better performance than other methods when using the IBN-Net50 as the backbone. For example, when compared with SNR, our method improve $+7.8\%$ ($74.3$ vs. $66.5$) on mAP. It is worth noting that, SNR uses the IN to remove the style information, thus this is a fair comparison when we use IBN-Net50 as the backbone. Moreover, compared with the baseline, using our method can achieve a great improvement. For example, on PRID, two baselines (\ie, IBN-Net50 and ResNet50) can be gained by $+21.6\%$ ($74.3$ vs. $52.7$) and $+11.8\%$ ($59.1$ vs. $47.3$) on mAP, thus this validates the efficacy of the proposed method when there are more source domains during the training course.

\begin{table}[htbp]
  \centering
  \caption{Experimental results on two small-scale datasets. The model is trained on five datasets including MARKET1501 (MA), DUKEMTMC-REID, CUHK03, CUHK02 and CUHK-SYSU. $^\dagger$ denotes that the model is trained using more data as in~\cite{DBLP:conf/cvpr/ChoiKJPK21}.}
    \begin{tabular}{l|cc|cc}
    \toprule
    \multicolumn{1}{c|}{\multirow{2}[1]{*}{Method}} & \multicolumn{2}{c|}{PRID} & \multicolumn{2}{c}{VIPeR} \\
\cmidrule{2-5}          & mAP   & Rank-1 & mAP   & Rank-1 \\
    \midrule
    Agg Align~\cite{zhang2017alignedreid} & 25.5  & 17.2  & 52.9  & 42.8  \\
    Reptile~\cite{nichol2018first} & 26.9  & 17.9  & 31.3  & 22.1  \\
    CrossGrad~\cite{DBLP:conf/iclr/ShankarPCCJS18} & 28.2  & 18.8  & 30.4  & 20.9  \\
    Agg PCB~\cite{DBLP:journals/pami/SunZLYTW21} & 32.0  & 21.5  & 45.4  & 38.1  \\
    MLDG~\cite{DBLP:conf/aaai/LiYSH18}  & 35.4  & 24.0  & 33.5  & 23.5  \\
    PPA~\cite{DBLP:conf/cvpr/QiaoLSY18}   & 45.3  & 31.9  & 54.5  & 45.1  \\
    DIMN~\cite{DBLP:conf/cvpr/SongYSXH19}  & 52.0  & 39.2  & 60.1  & 51.2  \\
    SNR~\cite{DBLP:conf/cvpr/JinLZ0Z20}   & 66.5  & 52.1  & 61.3  & 52.9  \\
    RaMoE~\cite{DBLP:conf/cvpr/DaiLLTD21} & 67.3  & 57.7  & 64.6  & \textbf{56.6} \\
    \midrule
    Baseline & 47.3  & 37.2  & 49.8  & 40.2  \\
    MixNorm \scriptsize{(ResNet-50)} & 59.1  & 49.2  & 60.6  & 50.8  \\
    \midrule
    Baseline & 52.7  & 42.6  & 57.9  & 47.7  \\
    MixNorm \scriptsize{(IBN-Net50)} & \textbf{74.3} & \textbf{65.2} & \textbf{66.6} & 56.4  \\
    \midrule
    MetaBIN$^\dagger$~\cite{DBLP:conf/cvpr/ChoiKJPK21}&  \textbf{81.0}  & \textbf{74.2} & 68.6  & 59.9\\
     MixNorm$^\dagger$ & 78.4  & 71.0  & \textbf{70.2} & \textbf{61.7}\\
    \bottomrule
    \end{tabular}%
  \label{tab01}%
\end{table}%

We perform the comparison between MetaBIN~\cite{DBLP:conf/cvpr/ChoiKJPK21} and our method. It is worth noting that MetaBIN utilize more data to train the model when compared with our method. For example, when using the Market1501 dataset during training, we only employ the training set of this dataset (\ie, 751 IDs). Differently, MetaBIN~\cite{DBLP:conf/cvpr/ChoiKJPK21} uses not only the training set but also query set and gallery set (\ie, 1501 IDs). Similarly, on other training set, MetaBIN also uses all samples (\ie, training set and testing set) to train the model. Besides, to conduct a fair comparison, we take IBN-Net50 (\ie, the normalization layer consists of instance normalization and batch normalization based on ResNet-50) as the backbone because MetaBIN also includes instance normalization and the batch normalization based on ResNet-50. In this experiment, we leverage the same training data to train our model as MetaBIN, and $^\dagger$ denotes that the model is trained using more data in Table~\ref{tab01}. As seen, when using more data to train the model, the performance of our method can be further increased. Besides, our method can obtain better performance  compared with MetaBIN on VIPeR. Moreover, despite our method is inferior to  MetaBIN on PRID, our method can indeed achieve a large improvement compared with the baseline.
\subsection{Ablation Studies} \label{sec:EXP-SS}
In this part, we conduct the ablation study to verify the effectiveness of each component in our method. Our method mainly consists of domain-aware mix-normalization (DMN) and domain-aware center regularization (DCR). Table~\ref{tab03} shows the experimental results when we employ ResNet-50 as the backbone. As seen, when adding the DMN into the baseline, the baseline can be gained in multiple different tasks. For example, in the ``D+C+Ms$\rightarrow$Ma'' task, using the DMN can increase the baseline by  $+8.1\%$ ($48.5$ vs. $40.4$) and $+7.3\%$ ($76.9$ vs. $69.6$) on mAP and Rank-1. This shows that using the DMN to enrich the diversity of the feature can indeed alleviate the model's overfitting and boost the generalization capability of the model to the unseen domain. Moreover, using DCR can further improve the model's robustness. For example, in the ``D+C+Ms$\rightarrow$Ma'' and ``Ma+D+Ms$\rightarrow$C'' tasks, adding the DCR can bring the  $+2.9\%$ ($51.4$ vs. $48.5$) and  $+2.3\%$ ($29.0$ vs. $26.7$) improvements on mAP, thus it confirms the effectiveness of DCR in our method, which can better utilize the diverse features from the DMN to learn the domain-invariant model.
\begin{table}[htbp]
  \centering
  \caption{Ablation study when ResNet-50 is utilized as the backbone.}
    \begin{tabular}{l|cccc}
    \toprule
    \multicolumn{1}{c|}{\multirow{2}[1]{*}{Module}} & mAP   & Rank-1 & Rank-5 & Rank-10 \\
\cmidrule{2-5}          & \multicolumn{4}{c}{D+C+Ms$\rightarrow$Ma} \\
    \midrule
    Baseline & 40.4  & 69.6  & 83.8  & 87.9  \\
    Baseline+DMN & 48.5  & 76.9  & 88.8  & 92.1  \\
    Baseline+DMN+DCR & \textbf{51.4} & \textbf{78.9} & \textbf{90.2} & \textbf{93.3} \\
    \midrule
          & \multicolumn{4}{c}{Ma+C+Ms$\rightarrow$D} \\
    \midrule
    Baseline & 44.2  & 64.0  & 77.0  & 81.5  \\
    Baseline+DMN & 48.4  & 67.9  & 80.4  & 83.7  \\
    Baseline+DMN+DCR & \textbf{49.9} & \textbf{70.8} & \textbf{81.7} & \textbf{85.0} \\
    \midrule
          & \multicolumn{4}{c}{Ma+D+Ms$\rightarrow$C} \\
    \midrule
    Baseline & 24.0  & 24.4  & 41.7  & 51.4  \\
    Baseline+DMN & 26.7  & 26.9  & 45.5  & 56.2  \\
    Baseline+DMN+DCR & \textbf{29.0} & \textbf{29.6} & \textbf{47.2} & \textbf{58.1} \\
    \bottomrule
    \end{tabular}%
  \label{tab03}%
\end{table}%

\begin{figure}
\centering
\subfigure[D+C+Ms$\rightarrow$Ma]{
\includegraphics[width=4.1cm]{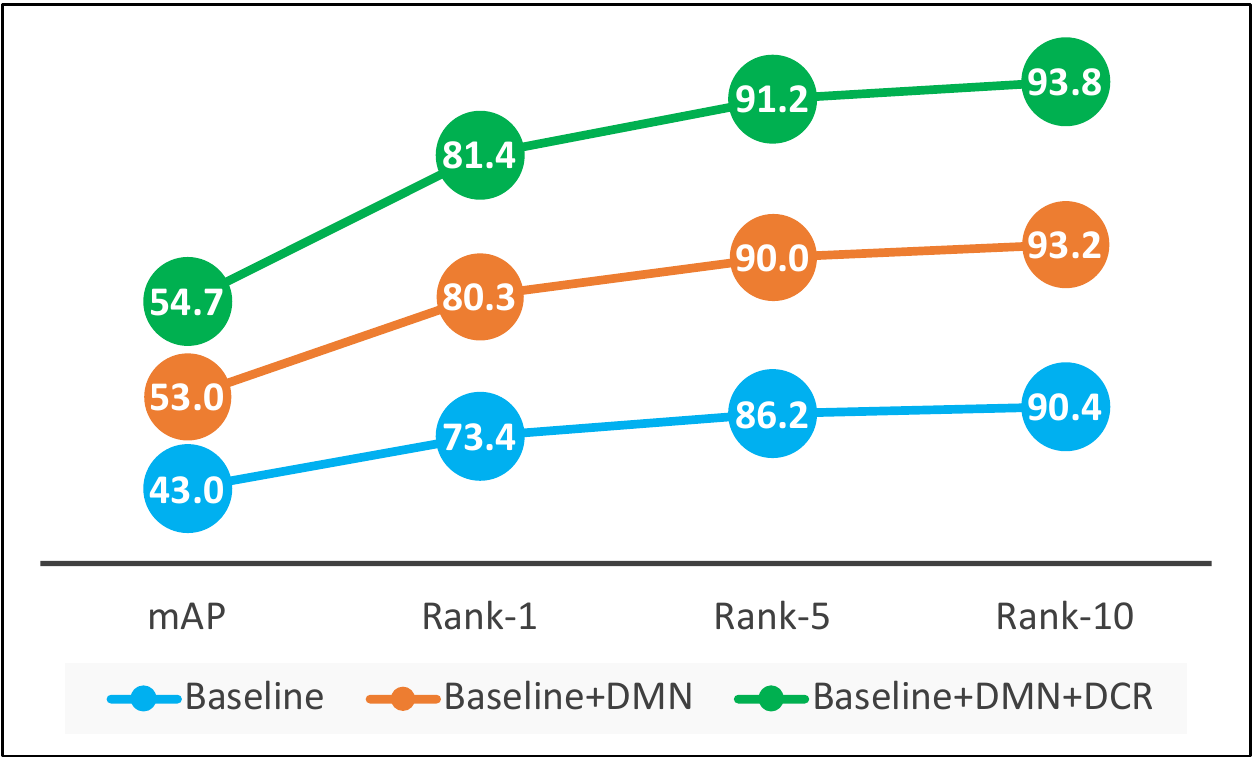}
}
\subfigure[Ma+D+Ms$\rightarrow$C]{
\includegraphics[width=4.03cm]{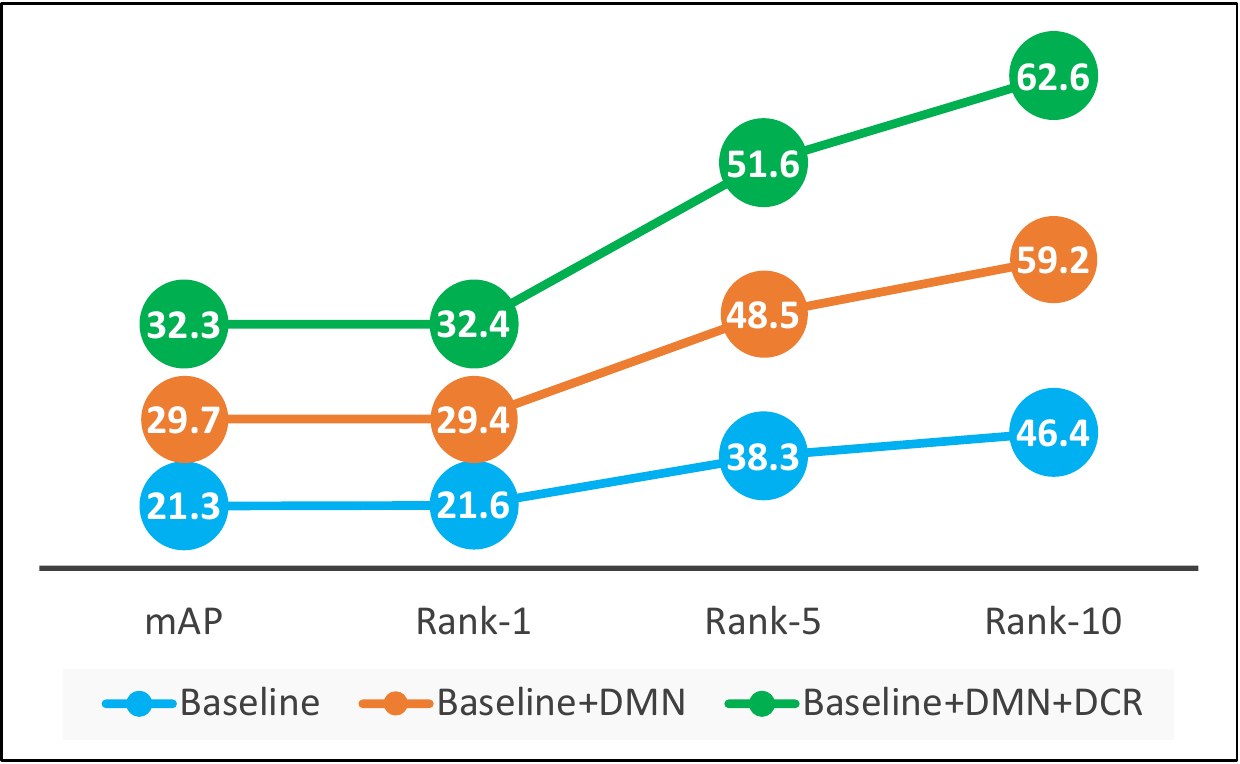}
}
\caption{Ablation study when IBN-Net50 is used as the backbone.}
\label{fig03}
\end{figure}

Besides, we also validate the efficacy of each component on IBN-Net50, as reported in Fig.~\ref{fig03}. As seen in all reported tasks, ``Baseline+DMN'' has better results than the ``Baseline''. Moreover, adding the DCR (\ie, ``Baseline+DMN+DCR'') can further improve the generalization capability to the unseen domains. For example, in the ``Ma+D+Ms$\rightarrow$C'' task, DMN can increase the baseline by  $+8.4\%$ ($29.7$ vs. $21.3$), while using the DCR can further gain $+2.6\%$ ($32.3$ vs. $29.7$) on mAP. This experiment deeply confirms the value of each component in our method.

\subsection{Further Analysis}\label{sec:EXP-FA}
\begin{figure}
\centering
\includegraphics[width=7cm]{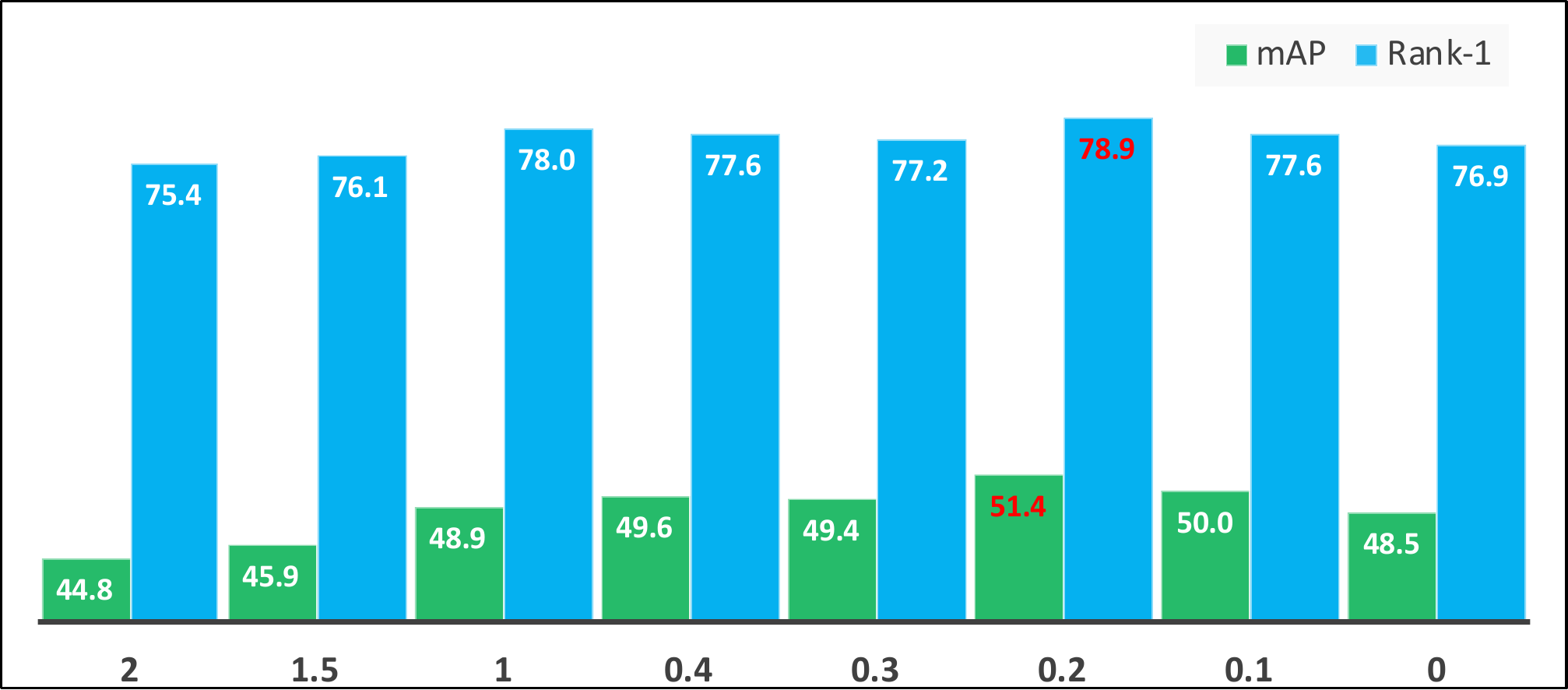}
\caption{The impact of $\lambda$ of Eq.~\ref{eq08} in the ``D+C+Ms$\rightarrow$Ma'' task. Note that the red is the best result.}
\label{fig01}
\end{figure}

\textbf{Hyper-parameter analysis.} We perform the experiment to evaluate the sensitivity of the hyper-parameter $\lambda$ in Eq.~\ref{eq08}. The experimental results are shown in Fig.~\ref{fig01}. As seen, when $\lambda$ is set as $0.2$, the best performance can be obtained. If using the smaller $\lambda$, the diverse features from the domain-aware mix-normalization could not be sufficiently exploited. If we use the larger $\lambda$, the model could more focus on reducing the domain gap across different source domains, thus it might slightly destroy the discriminative information of the learned features. According to the results in Fig.~\ref{fig01}, we set $\lambda$ as $0.2$ in all our experiments.

\begin{figure}[t]
\centering
\subfigure[D+C+Ms$\rightarrow$Ma]{
\includegraphics[width=4.1cm]{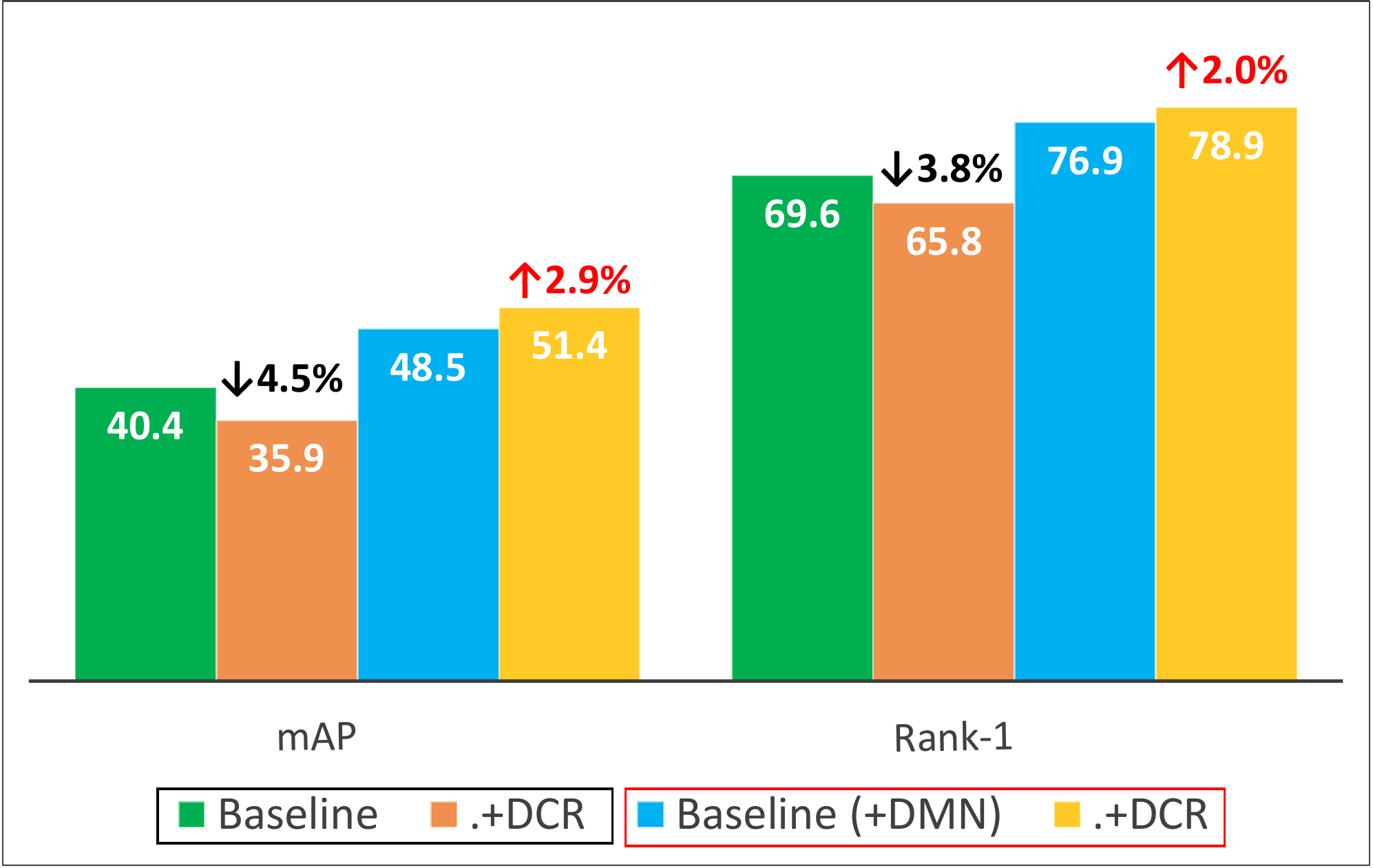}
}
\subfigure[Ma+C+Ms$\rightarrow$D]{
\includegraphics[width=4.1cm]{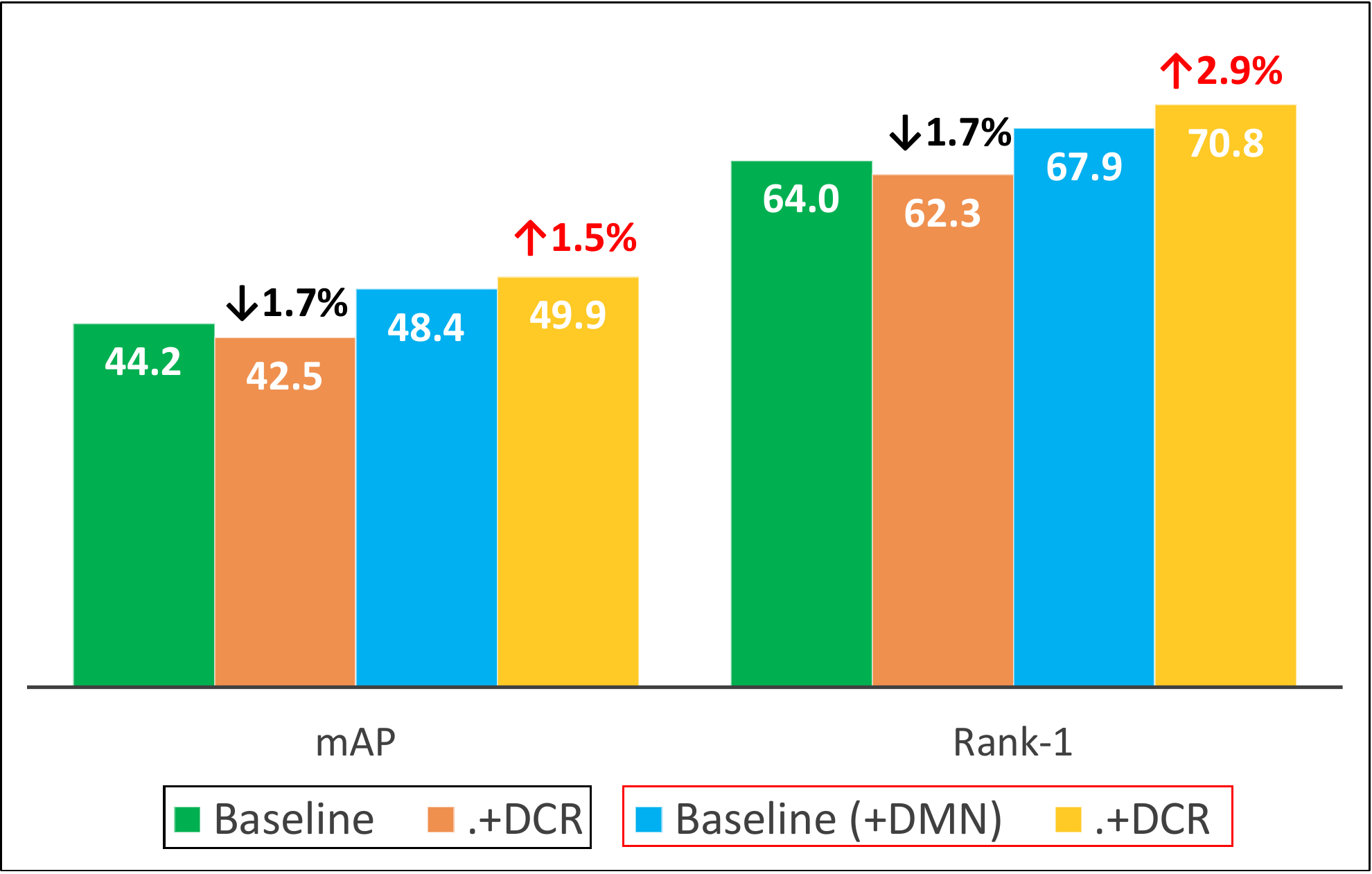}
}
\caption{The experimental results of domain-aware center regularization (DCR) used on different baselines in two tasks. Best viewed in color.}
\label{fig02}
\end{figure}

\textbf{More experiments on DCR.} Domain-aware center regularization (DCR) is utilized to better train the model when the DMN module produces diverse features. Here we also adopt it to the original baseline (\ie, the raw ResNet50), and the experimental results are displayed in Fig.~\ref{fig02}. As seen, if we directly use DCR on the original baseline, the performance will obviously decrease. For example, in the ``D+C+Ms$\rightarrow$Ma'' task, when leveraging the DCR on the baseline, the result will drop by $-4.5\%$ ($35.9$ vs. $40.4$) on mAP. Considering the pattern in the original baseline is fixed,  which does not need to focus on the domain gap, using the DCR could have a negative effect on the learning of the discriminative features. On the contrary, using the DCR on the baseline with DMN will obtain an obvious improvement.

\textbf{Comparison between DCR and center loss.} The center loss (CL) in~\cite{DBLP:conf/eccv/WenZL016} simultaneously learns a center for deep
features of each class and penalizes the distances between the deep features and their corresponding class centers, which can be also utilized to reduce the domain gap across different domains in the generalizable person Re-ID. In this experiment, we replace the domain-aware center regularization (DCR) by the center loss in our method. Similar to \cite{DBLP:conf/cvpr/DaiLLTD21}, we set $\lambda$ in Eq.~\ref{eq08} as $5\times 10^{-4}$ when using the center loss. The experimental results are reported in Table~\ref{tab05}. As seen, using the DCR can obtain better performance than CL. For example, in the ``Ma+C+Ms$\rightarrow$D'' task, the DCR exceeds CL by $+2.2\%$ ($49.9$ vs. $47.7$) on mAP. This mainly thanks to  that, this method aims to pull each domain close to the center of all domains in each batch, which can further promote the power of our DMN, thus it can better mitigate the domain gap across different domains.

\begin{table}[htbp]
  \centering
  \caption{Comparison between DCR and center loss (CL)~\cite{DBLP:conf/eccv/WenZL016}.}
    \begin{tabular}{l|cccc}
    \toprule
    \multicolumn{1}{c|}{\multirow{2}[1]{*}{Method}} & mAP   & Rank-1 & Rank-5 & Rank-10 \\
\cmidrule{2-5}          & \multicolumn{4}{c}{D+C+Ms$\rightarrow$Ma} \\
    \midrule
    CL    & 49.2  & 77.2  & 88.8  & 92.4 \\
    DCR & \textbf{51.4} & \textbf{78.9} & \textbf{90.2} & \textbf{93.3} \\
    \midrule
          & \multicolumn{4}{c}{Ma+C+Ms$\rightarrow$D} \\
          \midrule
    CL     & 47.7  & 67.7  & 79.6  & 83.7 \\
    DCR & \textbf{49.9} & \textbf{70.8} & \textbf{81.7} & \textbf{85.0} \\
    \midrule
          & \multicolumn{4}{c}{Ma+D+Ms$\rightarrow$C} \\
          \midrule
    CL    & 26.9  & 27.1  & 44.8  & 55.4 \\
    DCR & \textbf{29.0} & \textbf{29.6} & \textbf{47.2} & \textbf{58.1} \\
    \bottomrule
    \end{tabular}%
  \label{tab05}%
\end{table}%

\textbf{The effectiveness of the different domain combinations (DDC) for each DMN at each iteration.} During training, each proposed domain-aware mix-normalization (DMN) independently conducts the normalization operation, thus the domain's combination in Eqs.~\ref{eq05} and ~\ref{eq06} is different, \ie, the randomly selected domain set $\phi$ could be different in each DMN. Here we also conduct the experiment when using the same domain combination for all DMN at each iteration, as reported in Table~\ref{tab04}. As seen, the performance of using the same domain combination for all DMN at each iteration will decrease in general. For example, in the ``Ma+D+Ms$\rightarrow$C'' task, the result will drop by $-2.1\%$ ($26.9$ vs. $29.0$) and $-1.5\%$ ($28.1$ vs. $29.6$) on mAP and Rank-1. This main reason is that the randomization appears not only in each DMN but also inter the different DMNs when we adopt the independent DMN. Therefore, using the DDC scheme can further boost the diversity of features, so that it can enhance the generalization capability of the model to the unseen domain.

\begin{table}[htbp]
  \centering
  \caption{The experimental results of the different combinations (DDC) and the same domain combination for each DMN at each iteration during training}
    \begin{tabular}{c|cccc}
    \toprule
    \multicolumn{1}{c|}{Using DDC (Y/N)} & mAP   & Rank-1 & Rank-5 & Rank-10 \\
    \midrule
           & \multicolumn{4}{c}{D+C+Ms$\rightarrow$Ma} \\
    \midrule
         \xmark  & 49.2  & 77.6  & 88.4  & 92.2  \\
         $\checkmark$  & \textbf{51.4} & \textbf{78.9} & \textbf{90.2} & \textbf{93.3} \\
    \midrule
          & \multicolumn{4}{c}{Ma+C+Ms$\rightarrow$D} \\
    \midrule
         \xmark  & 49.2  & 68.2  & 81.0  & 83.9  \\
        $\checkmark$   & \textbf{49.9} & \textbf{70.8} & \textbf{81.7} & \textbf{85.0} \\
    \midrule
          & \multicolumn{4}{c}{Ma+D+C$\rightarrow$Ms} \\
    \midrule
         \xmark  & 19.1  & 46.3  & \textbf{60.2}  & \textbf{65.8}  \\
        $\checkmark$   & \textbf{19.4} & \textbf{47.2} & 60.0 & 65.4 \\
    \midrule
          & \multicolumn{4}{c}{Ma+D+Ms$\rightarrow$C} \\
    \midrule
         \xmark  & 26.9  & 28.1  & 45.1  & 54.3  \\
         $\checkmark$  & \textbf{29.0} & \textbf{29.6} & \textbf{47.2} & \textbf{58.1} \\
    \bottomrule
    \end{tabular}%
  \label{tab04}%
\end{table}%

\textbf{Analysis of the number of randomly selected domains in DMN.}
In our experiment, we observe that, if the maximum number of randomly selected domains in domain-aware mix-normalization (DMN) is equal to the number of all source domains, the performance will decrease, as shown in Table~\ref{tab06}. As observed, when setting the maximum domain number as $D$, the mAP and Rank-1,5,10 will slightly drop in the ``Ma+C+Ms$\rightarrow$D'' task. This could be because the final statistics for testing accumulate the statistics of all domains, thus if using the statistics of all domains to normalize the features in the training stage, the model could be mildly overfitting to source domains. Therefore, in all experiments, we set the maximum number of randomly selected domains as $D-1$.
\begin{table}[htbp]
  \centering
  \caption{Experimental results when the maximum number of randomly selected domains (MNRSD) in DMN is equal to the number of all source domains. Note that $D$ is the total number of source domains.}
    \begin{tabular}{l|rrrr}
    \toprule
    \multirow{2}[1]{*}{MNRSD} & \multicolumn{1}{c}{mAP} & \multicolumn{1}{c}{Rank-1} & \multicolumn{1}{c}{Rank-5} & \multicolumn{1}{c}{Rank-10} \\
\cmidrule{2-5}          & \multicolumn{4}{c}{D+C+Ms$\rightarrow$Ma} \\
    \midrule
    $D$ & 49.9  & 77.9  & 89.7  & 92.0  \\
    $D-1$  & \textbf{51.4} & \textbf{78.9} & \textbf{90.2} & \textbf{93.3} \\
    \midrule
          & \multicolumn{4}{c}{Ma+C+Ms$\rightarrow$D} \\
    \midrule
     $D$ & 49.3  & 69.1  & 80.6  & 84.9  \\
    $D-1$   & \textbf{49.9} & \textbf{70.8} & \textbf{81.7} & \textbf{85.0} \\
    \bottomrule
    \end{tabular}%
  \label{tab06}%
\end{table}%

\textbf{Analysis of different sampling schemes.}
Our method is based on the uniform sampling (US) scheme, \ie, we randomly select the same number of samples for all domains to generate each batch in the training stage. As mentioned in Section \ref{s-framework}, it could result in the model's overfitting to source domains due to the fixed pattern of the statistics. To further confirm the effectiveness of our method, we also perform the random sampling (RS) scheme during training, \ie, we first combine all domains into a set, and then randomly select IDs and samples to generate a batch. Particularly, the batch size of RS is the same as US, and all experimental setting is also consistent. The results are reported in Table~\ref{tab07}. As seen, compared with ``US(Baseline)'', ``RS(Baseline)'' can yield better results due to its randomness across different batches. However, the randomness is limited because the random sampling scheme still utilizes the same statistics to normalize all samples in a batch. On the contrary, our method randomly combines different domains to generate the local statistics to normalize the samples in each batch, which can produce diverse features to reduce the model's overfitting and enhance the generalization capability of the model to the unseen domain. As shown in Table~\ref{tab07}, our method (\ie, ``US(MixNorm)'') significantly outperforms the random sampling scheme in all tasks. Therefore, this experiment further verifies the effectiveness of our method.
\begin{table}[htbp]
  \centering
  \caption{Experimental results of different sampling methods. Note that RS and US indicate the random sampling and uniform sampling, respectively.}
    \begin{tabular}{l|cccc}
    \toprule
    \multicolumn{1}{c|}{\multirow{2}[1]{*}{Method}} & mAP   & Rank-1 & Rank-5 & Rank-10 \\
\cmidrule{2-5}          & \multicolumn{4}{c}{D+C+Ms$\rightarrow$Ma} \\
    \midrule
    RS (Baseline) & 44.5  & 72.9  & 85.8  & 89.6 \\
    US (Baseline) & 40.4  & 69.6  & 83.8  & 87.9 \\
    US (MixNorm) & \textbf{51.4} & \textbf{78.9} & \textbf{90.2} & \textbf{93.3} \\
    \midrule
          & \multicolumn{4}{c}{Ma+C+Ms$\rightarrow$D} \\
    \midrule
    RS (Baseline) & 46.1  & 65.2  & 79.1  & 83.0 \\
    US (Baseline) & 44.2  & 64.0  & 77.0  & 81.5 \\
    US (MixNorm) & \textbf{49.9} & \textbf{70.8} & \textbf{81.7} & \textbf{85.0} \\
    \midrule
          & \multicolumn{4}{c}{Ma+D+C$\rightarrow$Ms} \\
    \midrule
    RS (Baseline) & 16.1  & 41.2  & 54.9  & 61.2 \\
    US (Baseline) & 14.7  & 39.6  & 53.9  & 59.5 \\
    US (MixNorm) & \textbf{19.4} & \textbf{47.2} & \textbf{60.0} & \textbf{65.4} \\
    \midrule
          & \multicolumn{4}{c}{Ma+D+Ms$\rightarrow$C} \\
    \midrule
    RS (Baseline) & 26.0  & 26.2  & 43.6  & 54.1 \\
    US (Baseline) & 24.0  & 24.4  & 41.7  & 51.4 \\
    US (MixNorm) & \textbf{29.0} & \textbf{29.6} & \textbf{47.2} & \textbf{58.1} \\
    \bottomrule
    \end{tabular}%
  \label{tab07}%
\end{table}%

\textbf{Comparison between our MixNorm and MixStyle.} MixStyle~\cite{DBLP:conf/iclr/ZhouY0X21} is developed to tackle the domain generalization issue from the data augmentation in the feature-level, which mixes the style of different images to generate the style to produce the diverse features. We conduct a single experiment to compare it with our method, as displayed in Table~\ref{tab08}. In this experiment, we use the MixStyle with domain information to mix different styles, \ie, mixing samples from different domains to form the new style, and insert the module in the back of Block-1,2,3 of ResNet50. According to the experimental results in Table~\ref{tab08}, we observe that our MixNorm has better performance than MixStyle in all tasks, \eg, in the ``D+C+Ms$\rightarrow$Ma'' task, MixNorm exceeds the MixStyle by $+9.1\%$ ($51.4$ vs. $42.3$) on mAP. This mainly owes to the effectiveness of domain-aware mix-moralization and domain-aware center regularization in our MixNorm.

\begin{table}[htbp]
  \centering
  \caption{Comparison between our MixNorm and MixStyle~\cite{DBLP:conf/iclr/ZhouY0X21}.}
    \begin{tabular}{l|cccc}
    \toprule
    \multicolumn{1}{c|}{\multirow{2}[1]{*}{Method}} & mAP   & Rank-1 & Rank-5 & Rank-10 \\
\cmidrule{2-5}          & \multicolumn{4}{c}{D+C+Ms$\rightarrow$Ma} \\
    \midrule
    MixStyle~\cite{DBLP:conf/iclr/ZhouY0X21} & 42.3  & 71.5  & 84.9  & 89.1 \\
    MixNorm (ours) & \textbf{51.4} & \textbf{78.9} & \textbf{90.2} & \textbf{93.3} \\
    \midrule
          & \multicolumn{4}{c}{Ma+C+Ms$\rightarrow$D} \\
    \midrule
    MixStyle~\cite{DBLP:conf/iclr/ZhouY0X21} & 44.8  & 65.1  & 78.5  & 81.6 \\
    MixNorm (ours)& \textbf{49.9} & \textbf{70.8} & \textbf{81.7} & \textbf{85.0} \\
    \midrule
          & \multicolumn{4}{c}{Ma+D+C$\rightarrow$Ms} \\
    \midrule
    MixStyle~\cite{DBLP:conf/iclr/ZhouY0X21} & 16.4  & 42.2  & 55.9  & 61.7 \\
    MixNorm (ours)& \textbf{19.4} & \textbf{47.2} & \textbf{60.0} & \textbf{65.4} \\
    \midrule
          & \multicolumn{4}{c}{Ma+D+Ms$\rightarrow$C} \\
    \midrule
    MixStyle~\cite{DBLP:conf/iclr/ZhouY0X21} & 24.8  & 25.1  & 42.9  & 53.1 \\
    MixNorm (ours)& \textbf{29.0} & \textbf{29.6} & \textbf{47.2} & \textbf{58.1} \\
    \bottomrule
    \end{tabular}%
  \label{tab08}%
\end{table}%

\textbf{Evaluation on source domains.} In this part, to validate the effectiveness of our method on alleviating the overfitting issue, we compare our method with the baseline on the source domains, as shown in Table~\ref{tab09}. As seen, our method has poorer performance than the baseline on all source domains of all tasks. Moreover, we combine the above analysis and experiment (\ie, our method can significantly exceed the baseline on the unseen domain in all tasks), thus this confirms the proposed method can effectively mitigate the overfitting to the source domains. For example, in the ``D+Ms+C$\rightarrow$Ma'' task, the mAP of our method improves the baseline by $+11.0\%$ ($51.4$ vs. $40.4$) on the unseen domain (as shown in Table~\ref{tab03}), while the mAP of our method decreases by  $-4.2\%$ ($65.4$ vs. $69.6$) on the source domain ``DukeMTMC-reID''. Therefore, this validates that our method can mitigate the overfitting issue.
\begin{table}[htbp]
  \centering
  \caption{Experimental results on source domains.}
    \begin{tabular}{l|cc|cc|cc}
    \toprule
    \multicolumn{1}{c|}{Method} & mAP   & Rank-1 & mAP   & Rank-1 & mAP   & Rank-1 \\
    \midrule
          & \multicolumn{6}{c}{D+Ms+C$\rightarrow$Ma} \\
    \midrule
          & \multicolumn{2}{c|}{Test: D} & \multicolumn{2}{c|}{Test: Ms} & \multicolumn{2}{c}{Test: C} \\
    \midrule
    Baseline & 69.6  & 83.7  & 45.3  & 73.3 & 56.1  & 57.5 \\
    MixNorm & 65.4  & 81.0  & 43.1  & 73.1 & 45.5  & 45.3 \\
    \midrule
          & \multicolumn{6}{c}{Ma+Ms+C$\rightarrow$D} \\
    \midrule
          & \multicolumn{2}{c|}{Test: Ma} & \multicolumn{2}{c|}{Test: Ms} & \multicolumn{2}{c}{Test: C} \\
    \midrule
    Baseline & 79.6  & 92.2  & 45.5  & 73.5  & 58.8  & 60.9 \\
    MixNorm & 76.5  & 91.7  & 42.8  & 72.4  & 50.3  & 51.7 \\
    \midrule
          & \multicolumn{6}{c}{Ma+D+C$\rightarrow$Ms} \\
    \midrule
          & \multicolumn{2}{c|}{Test: Ma} & \multicolumn{2}{c|}{Test: D} & \multicolumn{2}{c}{Test: C} \\
    \midrule
    Baseline & 79.8  & 92.2  & 69.0  & 83.8  & 58.2  & 60.4 \\
    MixNorm & 76.9  & 92.0  & 66.1  & 82.5  & 48.3  & 48.8 \\
    \midrule
          & \multicolumn{6}{c}{Ma+D+Ms$\rightarrow$C} \\
    \midrule
          & \multicolumn{2}{c|}{Test: Ma} & \multicolumn{2}{c|}{Test: D} & \multicolumn{2}{c}{Test: Ms} \\
    \midrule
    Baseline & 80.4  & 92.1  & 71.8  & 84.9  & 47.2  & 74.7 \\
    MixNorm & 76.6  & 91.5  & 67.4  & 82.4  & 42.9  & 72.1 \\
    \bottomrule
    \end{tabular}%
  \label{tab09}%
\end{table}%

\begin{figure}
\centering

\subfigure[Baseline (D+Ms+C)]{
\includegraphics[width=4cm]{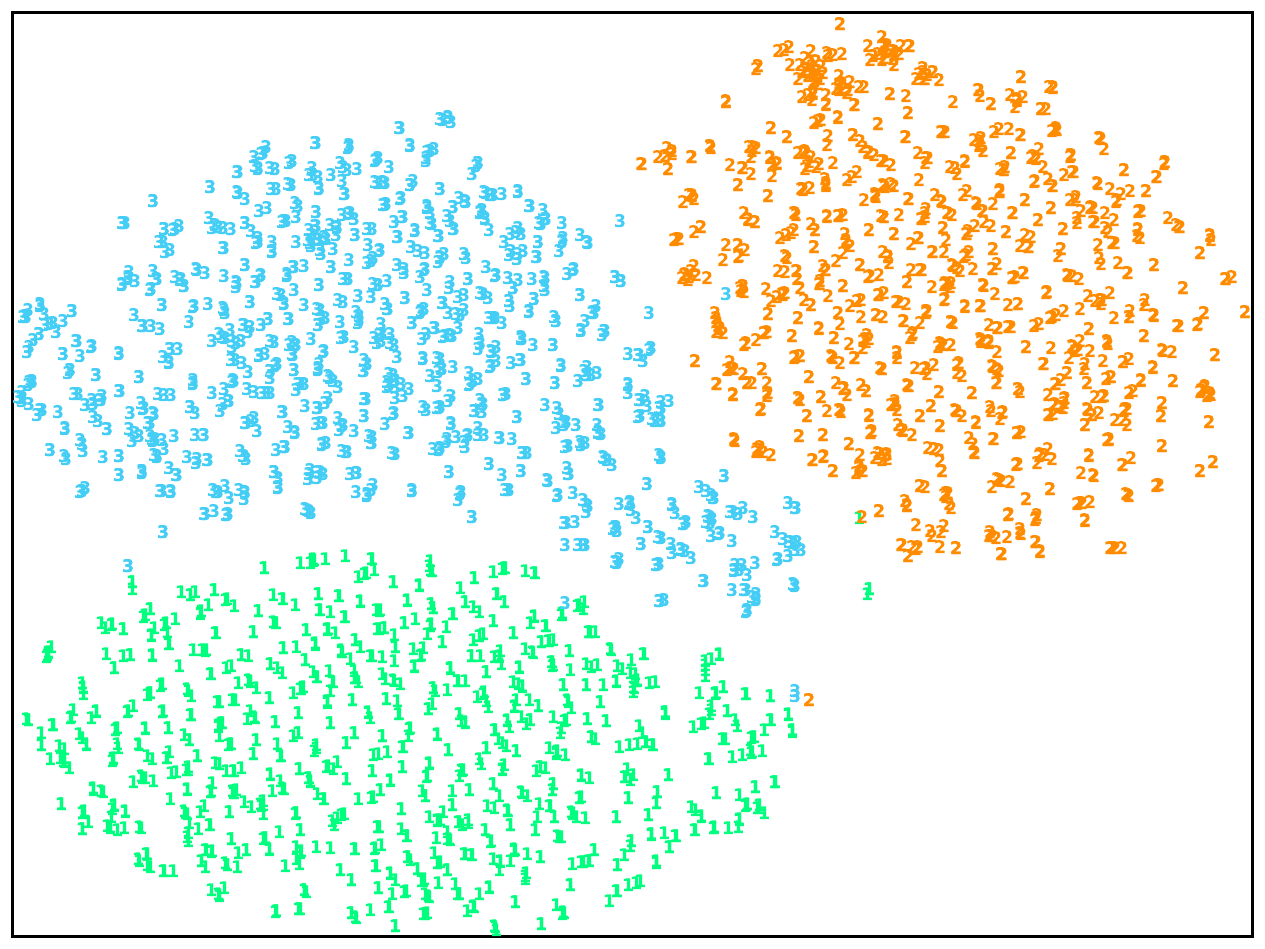}
}
\subfigure[MixNorm (D+Ms+C)]{
\includegraphics[width=4cm]{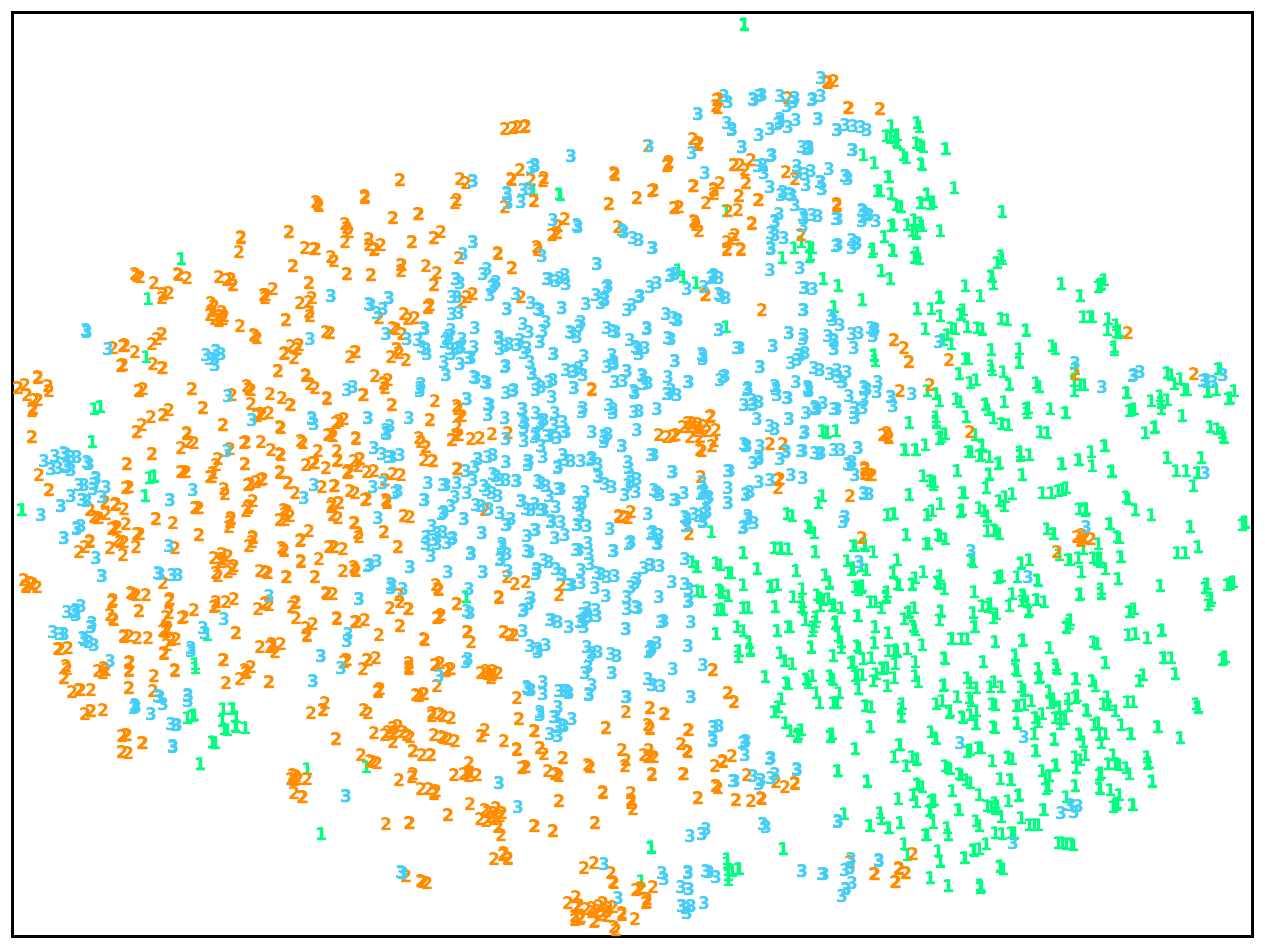}
}
\subfigure[Baseline (Ma+Ms+C)]{
\includegraphics[width=4cm]{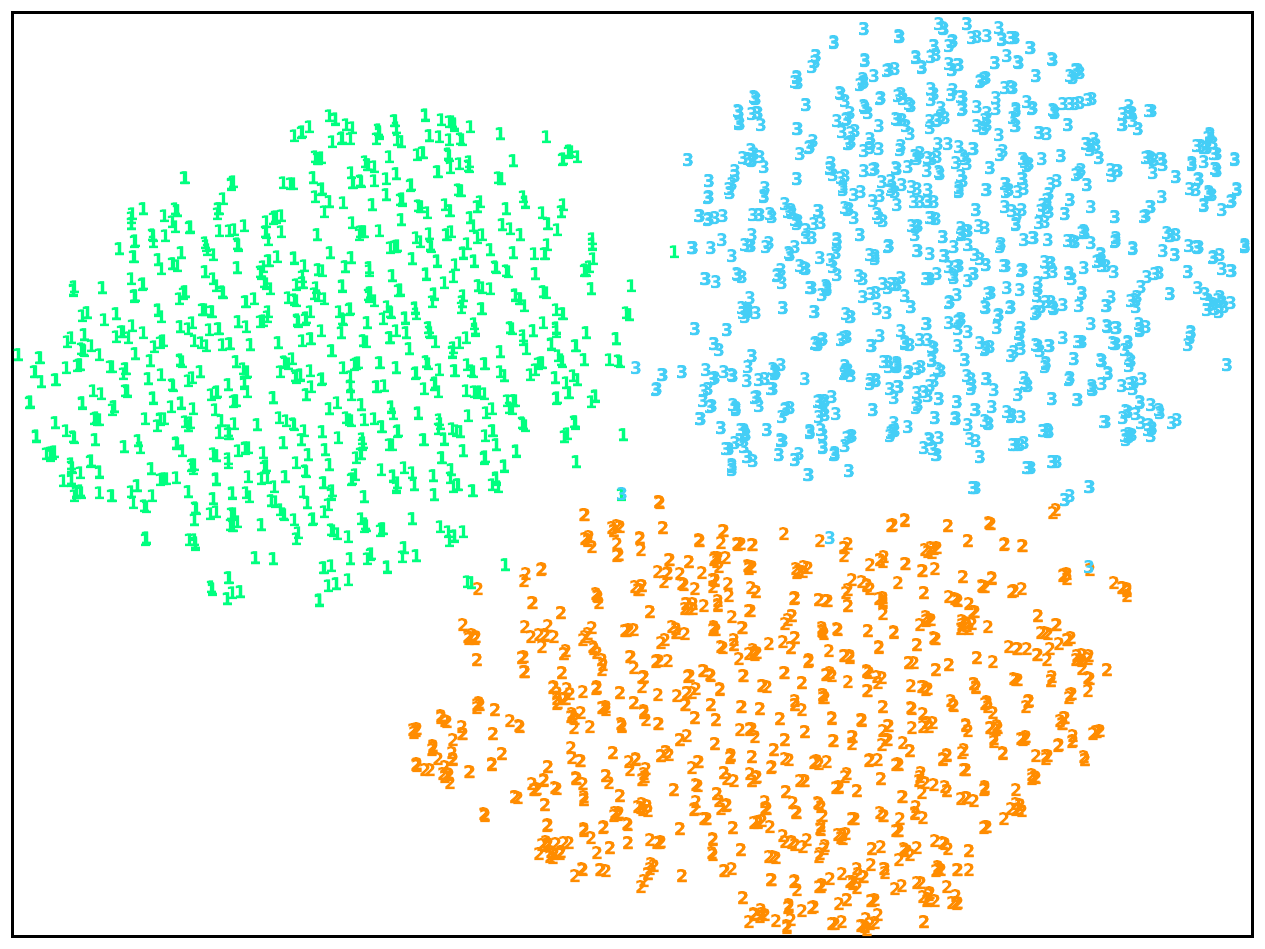}
}
\subfigure[MixNorm (Ma+Ms+C)]{
\includegraphics[width=4cm]{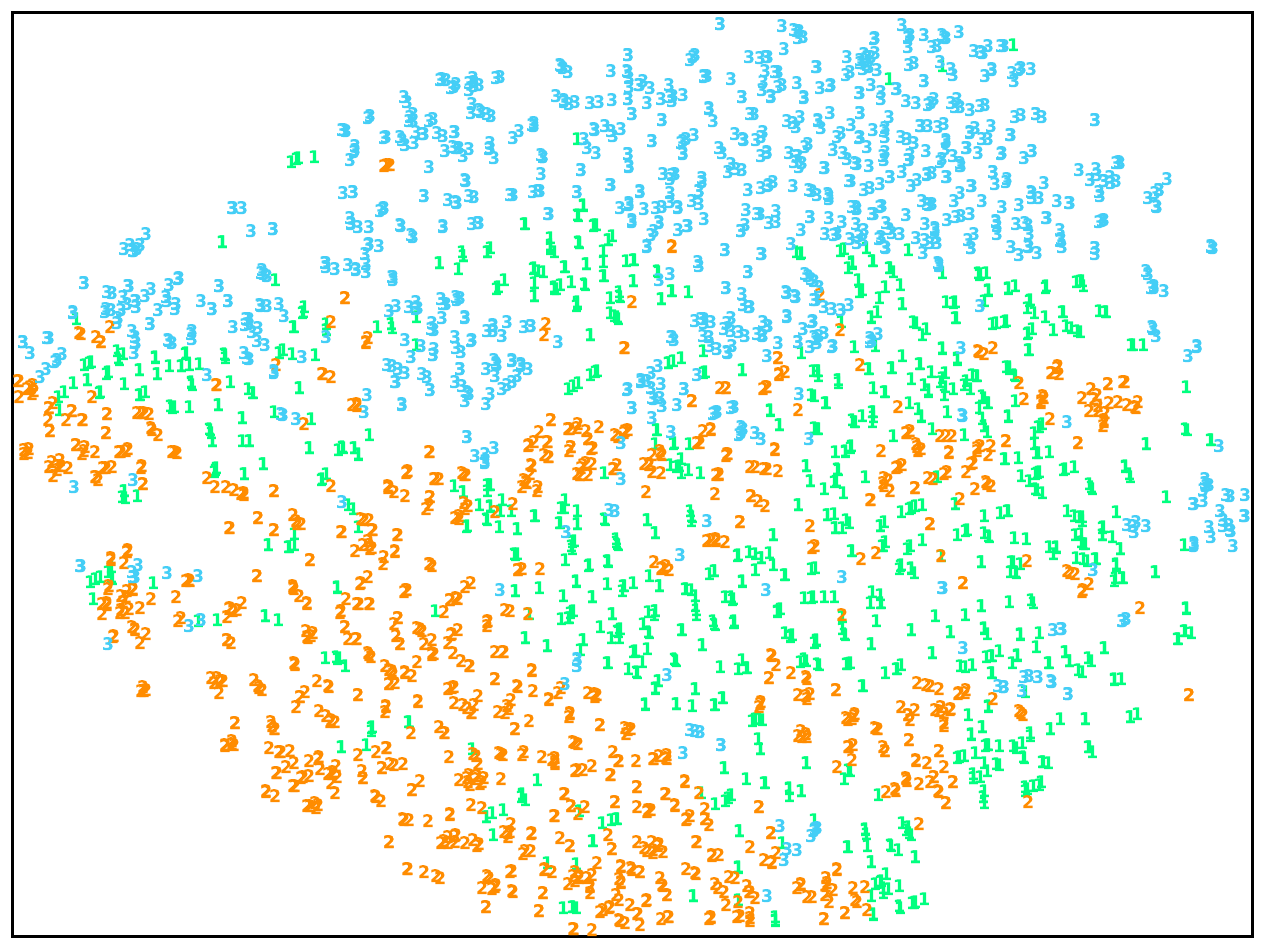}
}

\caption{Visualization of the features via t-SNE~\cite{van2008visualizing} on source domains. (a) and (b) are the visualization of feature representations from the model trained on DukeMTMC-reID, MSMT17 and CUHK03, and (c) and (d) are the visualization of feature representations from the model trained on Market1501, MSMT17 and CUHK03. Different colors denote different domains.}
\label{fig05}
\end{figure}

\textbf{Visualization of the feature distribution.} In Fig.~\ref{fig05}, we visualize the feature distribution on source domains. As observed in this figure, the features of different domains from the baseline are scattered in different regions, while our method obviously tends to mix all the source domains into the same region, which validates that our method can well learn the domain-invariant features so as to capture the generalizable model to the unseen domain.

\textbf{Evaluation on DMN in different layers.} In this part, we perform the experiment when using our DMN in different layers as reported in Table~\ref{tab_22}. In this table, ``All'' indicates replacing all BNs with DMN. As observed in this table,  using the DMN in each layer can bring the improvement compared with the baseline. Particularly, when we use DMN  in all layers, the best result can be obtained. 

\begin{table}[htbp]
  \centering
  \caption{The experimental of using the proposed DMN in different layers.}
  \label{tab_22}%
    \begin{tabular}{l|cccc}
    \toprule
    \multicolumn{1}{c|}{\multirow{2}[1]{*}{Layer}} & mAP   & Rank-1 & Rank-5 & Rank-10 \\
\cmidrule{2-5}          & \multicolumn{4}{c}{D+C+Ms$\rightarrow$Ma} \\
    \midrule
    Baseline & 40.4  & 69.6  & 83.8  & 87.9 \\
    \midrule
    layer-1 & 44.1  & 72.2  & 85.7  & 90.1 \\
    layer-2 & 43.6  & 72.9  & 86.5  & 90.3 \\
    layer-3 & 46.0    & 75.2  & 87.7  & 91.2 \\
    layer-4 & 46.6  & 73.8  & 86.8  & 90.5 \\
    \midrule
    All (ours) & \textbf{51.4} & \textbf{78.9} & \textbf{90.2} & \textbf{93.3} \\
    \midrule
          & \multicolumn{4}{c}{Ma+C+Ms$\rightarrow$D} \\
    \midrule
    Baseline & 44.2  & 64.0  & 77.0  & 81.5 \\
    \midrule
    layer-1 & 47.7  & 68.2  & 79.4  & 83.6 \\
    layer-2 & 46.5  & 66.7  & 80.3  & 83.6 \\
    layer-3 & 46.6  & 66.9  & 79.3  & 84.2 \\
    layer-4 & 47.6  & 68.1  & 79.3  & 83.4 \\
    \midrule
    All (ours) & \textbf{49.9} & \textbf{70.8} & \textbf{81.7} & \textbf{85.0} \\
    \midrule
          & \multicolumn{4}{c}{Ma+D+C$\rightarrow$Ms} \\
    \midrule
    Baseline & 14.7  & 39.6  & 53.9  & 59.5 \\
    \midrule
    layer-1 & 16.3  & 42.4  & 55.2  & 60.9 \\
    layer-2 & 16.6  & 42.3  & 55.9  & 61.6 \\
    layer-3 & 16.6  & 42.9  & 56.2  & 62.1 \\
    layer-4 & 16.1  & 42.7  & 56.2  & 61.7 \\
    \midrule
    All (ours) & \textbf{19.4} & \textbf{47.2} & \textbf{60.0} & \textbf{65.4} \\
    \bottomrule
    \end{tabular}%
\end{table}%

\textbf{Further analysis of DMN.}
 To further analyze the property of DMN, we always set $C$ in Algorithm~\ref{al01} as $1$ as displayed in Table~\ref{tab_r03}. As observed in this table, when $C$ is set as $1$, the proposed method can outperform the baseline clearly, which is attributed to the diverse statistics in a batch (\ie, each domain is normalized by the the statistics of the own domain instead of shared statistics of all domains). Besides, if by randomly combining different domains to normalize themselves, our method can achieve the better performance because more diverse features are generated.
 
\begin{table}[htbp]
  \centering
  \caption{The experimental results with different $C$ in algorithm~\ref{alg01} of our main paper. }
    \label{tab_r03}%
    \begin{tabular}{l|rrrr}
    \toprule
    \multicolumn{1}{c|}{\multirow{2}[1]{*}{$C$}} & \multicolumn{1}{c}{mAP} & \multicolumn{1}{c}{Rank-1} & \multicolumn{1}{c}{Rank-5} & \multicolumn{1}{c}{Rank-10} \\
\cmidrule{2-5}          & \multicolumn{4}{c}{D+C+Ms$\rightarrow$Ma} \\
    \midrule
     Baseline & 40.4  & 69.6  & 83.8  & 87.9 \\
    \midrule
    1     & 48.4  & 76.4  & 89.2  & 92.8  \\
    Ours     & \textbf{51.4} & \textbf{78.9} & \textbf{90.2} & \textbf{93.3} \\
    \midrule
          & \multicolumn{4}{c}{Ma+C+Ms$\rightarrow$D} \\
    \midrule
     Baseline & 44.2  & 64.0  & 77.0  & 81.5 \\
       \midrule
    1     & 47.0  & 67.2  & 79.2  & 83.0  \\
    Ours     & \textbf{49.9} & \textbf{70.8} & \textbf{81.7} & \textbf{85.0} \\
    \bottomrule
    \end{tabular}%
\end{table}%

\section{Conclusion}\label{s-conclusion}
In this paper, we aim to tackle the generalizable multi-source person Re-ID task from the data augmentation view. Different from the existing methods, we propose a novel domain-aware mix-normalization (DMN) method to achieve data augmentation in the feature level, which can generate diverse features to prevent the overfitting of the model to source domains. Furthermore, we employ a domain-aware center regularization (DCR) for better using the diverse features from DMN, which can map all features into the same space, thus it can enforce to learn the domain-invariant feature representations. We conduct the experiment on multiple datasets, which demonstrates the effectiveness of the proposed method.


%
%

\ifCLASSOPTIONcaptionsoff
  \newpage
\fi

\bibliographystyle{IEEEtran}
\bibliography{sigproc}

\begin{thebibliography}{10}
\providecommand{\url}[1]{#1}
\csname url@samestyle\endcsname
\providecommand{\newblock}{\relax}
\providecommand{\bibinfo}[2]{#2}
\providecommand{\BIBentrySTDinterwordspacing}{\spaceskip=0pt\relax}
\providecommand{\BIBentryALTinterwordstretchfactor}{4}
\providecommand{\BIBentryALTinterwordspacing}{\spaceskip=\fontdimen2\font plus
\BIBentryALTinterwordstretchfactor\fontdimen3\font minus
  \fontdimen4\font\relax}
\providecommand{\BIBforeignlanguage}[2]{{%
\expandafter\ifx\csname l@#1\endcsname\relax
\typeout{** WARNING: IEEEtran.bst: No hyphenation pattern has been}%
\typeout{** loaded for the language `#1'. Using the pattern for}%
\typeout{** the default language instead.}%
\else
\language=\csname l@#1\endcsname
\fi
#2}}
\providecommand{\BIBdecl}{\relax}
\BIBdecl

\bibitem{zheng2016person}
L.~Zheng, Y.~Yang, and A.~G. Hauptmann, ``Person re-identification: Past,
  present and future,'' \emph{arXiv preprint arXiv:1610.02984}, 2016.

\bibitem{ye2020deep}
M.~Ye, J.~Shen, G.~Lin, T.~Xiang, L.~Shao, and S.~C. Hoi, ``Deep learning for
  person re-identification: A survey and outlook,'' \emph{arXiv preprint
  arXiv:2001.04193}, 2020.

\bibitem{DBLP:journals/tcsv/LengYT20}
Q.~Leng, M.~Ye, and Q.~Tian, ``A survey of open-world person
  re-identification,'' \emph{IEEE Transactions on Circuits and Systems for
  Video Technology (TCSVT)}, vol.~30, no.~4, pp. 1092--1108, 2020.

\bibitem{DBLP:journals/pami/LiZG20}
M.~Li, X.~Zhu, and S.~Gong, ``Unsupervised tracklet person re-identification,''
  \emph{IEEE Transactions on Pattern Analysis and Machine Intelligence
  (TPAMI)}, vol.~42, no.~7, pp. 1770--1782, 2020.

\bibitem{chen2021occlude}
P.~Chen, W.~Liu, P.~Dai, J.~Liu, Q.~Ye, M.~Xu, Q.~Chen, and R.~Ji, ``Occlude
  them all: Occlusion-aware attention network for occluded person re-id,'' in
  \emph{International Conference on Computer Vision (ICCV)}, 2021, pp.
  11\,833--11\,842.

\bibitem{DBLP:journals/tcsv/QiWHSG20}
L.~Qi, L.~Wang, J.~Huo, Y.~Shi, and Y.~Gao, ``Progressive cross-camera
  soft-label learning for semi-supervised person re-identification,''
  \emph{IEEE Transactions on Circuits and Systems for Video Technology
  (TCSVT)}, vol.~30, no.~9, pp. 2815--2829, 2020.

\bibitem{DBLP:journals/tmm/ZhaoLZZWM20}
C.~Zhao, X.~Lv, Z.~Zhang, W.~Zuo, J.~Wu, and D.~Miao, ``Deep fusion feature
  representation learning with hard mining center-triplet loss for person
  re-identification,'' \emph{IEEE Transactions on Multimedia (TMM)}, vol.~22,
  no.~12, pp. 3180--3195, 2020.

\bibitem{DBLP:journals/tmm/WeiZY0019}
L.~Wei, S.~Zhang, H.~Yao, W.~Gao, and Q.~Tian, ``{GLAD:} global-local-alignment
  descriptor for scalable person re-identification,'' \emph{IEEE Transactions
  on Multimedia (TMM)}, vol.~21, no.~4, pp. 986--999, 2019.

\bibitem{DBLP:conf/cvpr/WuZGL19}
A.~Wu, W.~Zheng, X.~Guo, and J.~Lai, ``Distilled person re-identification:
  Towards a more scalable system,'' in \emph{IEEE Conference on Computer Vision
  and Pattern Recognition (CVPR)}, 2019, pp. 1187--1196.

\bibitem{DBLP:conf/cvpr/LiZG18}
W.~Li, X.~Zhu, and S.~Gong, ``Harmonious attention network for person
  re-identification,'' in \emph{IEEE Conference on Computer Vision and Pattern
  Recognition (CVPR)}, 2018, pp. 2285--2294.

\bibitem{DBLP:conf/cvpr/ZhengYY00K19}
Z.~Zheng, X.~Yang, Z.~Yu, L.~Zheng, Y.~Yang, and J.~Kautz, ``Joint
  discriminative and generative learning for person re-identification,'' in
  \emph{IEEE Conference on Computer Vision and Pattern Recognition (CVPR)},
  2019, pp. 2138--2147.

\bibitem{DBLP:journals/tomccap/QiWHSG21}
L.~Qi, L.~Wang, J.~Huo, Y.~Shi, and Y.~Gao, ``Greyreid: {A} novel two-stream
  deep framework with rgb-grey information for person re-identification,''
  \emph{ACM Transactions on Multimedia Computing, Communications, and
  Applications (TOMM)}, vol.~17, no.~1, pp. 27:1--27:22, 2021.

\bibitem{hou2020iaunet}
R.~Hou, B.~Ma, H.~Chang, X.~Gu, S.~Shan, and X.~Chen, ``Iaunet: Global
  context-aware feature learning for person reidentification,'' \emph{IEEE
  Transactions on Neural Networks and Learning Systems (TNNLS)}, vol.~32,
  no.~10, pp. 4460--4474, 2020.

\bibitem{hou2021feature}
{R. Hou, B. Ma, H. Chang, X. Gu, S. Shan, and X. Chen}, ``Feature completion
  for occluded person re-identification,'' \emph{IEEE Transactions on Pattern
  Analysis and Machine Intelligence (TPAMI)}, 2021.

\bibitem{DBLP:journals/tmm/YangYLJXYGHG21}
F.~Yang, K.~Yan, S.~Lu, H.~Jia, D.~Xie, Z.~Yu, X.~Guo, F.~Huang, and W.~Gao,
  ``Part-aware progressive unsupervised domain adaptation for person
  re-identification,'' \emph{IEEE Transactions on Multimedia (TMM)}, vol.~23,
  pp. 1681--1695, 2021.

\bibitem{DBLP:conf/cvpr/ZhaiLYSCJ020}
Y.~Zhai, S.~Lu, Q.~Ye, X.~Shan, J.~Chen, R.~Ji, and Y.~Tian, ``Ad-cluster:
  Augmented discriminative clustering for domain adaptive person
  re-identification,'' in \emph{IEEE Conference on Computer Vision and Pattern
  Recognition (CVPR)}, 2020, pp. 9018--9027.

\bibitem{DBLP:conf/iccv/WuZL19}
A.~Wu, W.~Zheng, and J.~Lai, ``Unsupervised person re-identification by
  camera-aware similarity consistency learning,'' in \emph{International
  Conference on Computer Vision (ICCV)}, 2019, pp. 6921--6930.

\bibitem{DBLP:conf/eccv/ChenLL020}
G.~Chen, Y.~Lu, J.~Lu, and J.~Zhou, ``Deep credible metric learning for
  unsupervised domain adaptation person re-identification,'' in \emph{European
  Conference on Computer Vision (ECCV)}, 2020, pp. 643--659.

\bibitem{DBLP:conf/iccv/QiWHZSG19}
L.~Qi, L.~Wang, J.~Huo, L.~Zhou, Y.~Shi, and Y.~Gao, ``A novel unsupervised
  camera-aware domain adaptation framework for person re-identification,'' in
  \emph{International Conference on Computer Vision (ICCV)}, 2019, pp.
  8079--8088.

\bibitem{zhou2021domain}
K.~Zhou, Z.~Liu, Y.~Qiao, T.~Xiang, and C.~C. Loy, ``Domain generalization: A
  survey,'' \emph{arXiv preprint arXiv:2103.02503}, 2021.

\bibitem{DBLP:conf/eccv/LiaoS20}
S.~Liao and L.~Shao, ``Interpretable and generalizable person re-identification
  with query-adaptive convolution and temporal lifting,'' in \emph{European
  Conference on Computer Vision (ECCV)}, 2020, pp. 456--474.

\bibitem{DBLP:conf/cvpr/ZhaoZYLLLS21}
Y.~Zhao, Z.~Zhong, F.~Yang, Z.~Luo, Y.~Lin, S.~Li, and N.~Sebe, ``Learning to
  generalize unseen domains via memory-based multi-source meta-learning for
  person re-identification,'' in \emph{IEEE Conference on Computer Vision and
  Pattern Recognition (CVPR)}, 2021, pp. 6277--6286.

\bibitem{DBLP:conf/cvpr/DaiLLTD21}
Y.~Dai, X.~Li, J.~Liu, Z.~Tong, and L.~Duan, ``Generalizable person
  re-identification with relevance-aware mixture of experts,'' in \emph{IEEE
  Conference on Computer Vision and Pattern Recognition (CVPR)}, 2021, pp.
  16\,145--16\,154.

\bibitem{DBLP:conf/icml/IoffeS15}
S.~Ioffe and C.~Szegedy, ``Batch normalization: Accelerating deep network
  training by reducing internal covariate shift,'' in \emph{International
  Conference on Machine Learning (ICML)}, 2015, pp. 448--456.

\bibitem{DBLP:conf/cvpr/GuoZZCLL20}
J.~Guo, X.~Zhu, C.~Zhao, D.~Cao, Z.~Lei, and S.~Z. Li, ``Learning meta face
  recognition in unseen domains,'' in \emph{IEEE Conference on Computer Vision
  and Pattern Recognition (CVPR)}, 2020, pp. 6162--6171.

\bibitem{ulyanov2016instance}
D.~Ulyanov, A.~Vedaldi, and V.~Lempitsky, ``Instance normalization: The missing
  ingredient for fast stylization,'' \emph{arXiv preprint arXiv:1607.08022},
  2016.

\bibitem{DBLP:conf/bmvc/JiaRH19}
J.~Jia, Q.~Ruan, and T.~M. Hospedales, ``Frustratingly easy person
  re-identification: Generalizing person re-id in practice,'' in \emph{British
  Machine Vision Conference (BMVC)}, 2019, p. 117.

\bibitem{DBLP:conf/cvpr/JinLZ0Z20}
X.~Jin, C.~Lan, W.~Zeng, Z.~Chen, and L.~Zhang, ``Style normalization and
  restitution for generalizable person re-identification,'' in \emph{IEEE
  Conference on Computer Vision and Pattern Recognition (CVPR)}, 2020, pp.
  3140--3149.

\bibitem{DBLP:conf/cvpr/SongYSXH19}
J.~Song, Y.~Yang, Y.~Song, T.~Xiang, and T.~M. Hospedales, ``Generalizable
  person re-identification by domain-invariant mapping network,'' in \emph{IEEE
  Conference on Computer Vision and Pattern Recognition (CVPR)}, 2019, pp.
  719--728.

\bibitem{DBLP:conf/cvpr/ChoiKJPK21}
S.~Choi, T.~Kim, M.~Jeong, H.~Park, and C.~Kim, ``Meta batch-instance
  normalization for generalizable person re-identification,'' in \emph{IEEE
  Conference on Computer Vision and Pattern Recognition (CVPR)}, 2021, pp.
  3425--3435.

\bibitem{DBLP:conf/cvpr/NamLPYY21}
H.~Nam, H.~Lee, J.~Park, W.~Yoon, and D.~Yoo, ``Reducing domain gap by reducing
  style bias,'' in \emph{IEEE Conference on Computer Vision and Pattern
  Recognition (CVPR)}, 2021, pp. 8690--8699.

\bibitem{DBLP:conf/eccv/SeoSKKHH20}
S.~Seo, Y.~Suh, D.~Kim, G.~Kim, J.~Han, and B.~Han, ``Learning to optimize
  domain specific normalization for domain generalization,'' in \emph{European
  Conference on Computer Vision (ECCV)}, 2020, pp. 68--83.

\bibitem{DBLP:conf/iccv/YueZZSKG19}
X.~Yue, Y.~Zhang, S.~Zhao, A.~L. Sangiovanni{-}Vincentelli, K.~Keutzer, and
  B.~Gong, ``Domain randomization and pyramid consistency: Simulation-to-real
  generalization without accessing target domain data,'' in \emph{International
  Conference on Computer Vision (ICCV)}, 2019, pp. 2100--2110.

\bibitem{DBLP:conf/cvpr/CarlucciDBCT19}
F.~M. Carlucci, A.~D'Innocente, S.~Bucci, B.~Caputo, and T.~Tommasi, ``Domain
  generalization by solving jigsaw puzzles,'' in \emph{IEEE Conference on
  Computer Vision and Pattern Recognition (CVPR)}, 2019, pp. 2229--2238.

\bibitem{DBLP:conf/nips/BalajiSC18}
Y.~Balaji, S.~Sankaranarayanan, and R.~Chellappa, ``Metareg: Towards domain
  generalization using meta-regularization,'' in \emph{Advances in Neural
  Information Processing Systems (NeurIPS)}, 2018, pp. 1006--1016.

\bibitem{DBLP:conf/iccv/LiZYLSH19}
D.~Li, J.~Zhang, Y.~Yang, C.~Liu, Y.~Song, and T.~M. Hospedales, ``Episodic
  training for domain generalization,'' in \emph{International Conference on
  Computer Vision (ICCV)}, 2019, pp. 1446--1455.

\bibitem{DBLP:conf/eccv/LiTGLLZT18}
Y.~Li, X.~Tian, M.~Gong, Y.~Liu, T.~Liu, K.~Zhang, and D.~Tao, ``Deep domain
  generalization via conditional invariant adversarial networks,'' in
  \emph{European Conference on Computer Vision (ECCV)}, 2018, pp. 647--663.

\bibitem{DBLP:journals/pr/ZhangQSG22}
J.~Zhang, L.~Qi, Y.~Shi, and Y.~Gao, ``Generalizable model-agnostic semantic
  segmentation via target-specific normalization,'' \emph{Pattern Recognition
  (PR)}, vol. 122, p. 108292, 2022.

\bibitem{DBLP:conf/eccv/ZhouYHX20}
{K. Zhou, Y. Yang, T. M. Hospedales, and T. Xiang}, ``Learning to generate
  novel domains for domain generalization,'' in \emph{European Conference on
  Computer Vision (ECCV)}, 2020, pp. 561--578.

\bibitem{DBLP:conf/aaai/ZhouYHX20}
K.~Zhou, Y.~Yang, T.~M. Hospedales, and T.~Xiang, ``Deep domain-adversarial
  image generation for domain generalisation,'' in \emph{AAAI Conference on
  Artificial Intelligence (AAAI)}, 2020, pp. 13\,025--13\,032.

\bibitem{DBLP:conf/cvpr/XuZ0W021}
Q.~Xu, R.~Zhang, Y.~Zhang, Y.~Wang, and Q.~Tian, ``A fourier-based framework
  for domain generalization,'' in \emph{IEEE Conference on Computer Vision and
  Pattern Recognition (CVPR)}, 2021, pp. 14\,383--14\,392.

\bibitem{DBLP:conf/eccv/HuangWXH20}
Z.~Huang, H.~Wang, E.~P. Xing, and D.~Huang, ``Self-challenging improves
  cross-domain generalization,'' in \emph{European Conference on Computer
  Vision (ECCV)}, 2020, pp. 124--140.

\bibitem{Li_2021_ICCV}
P.~Li, D.~Li, W.~Li, S.~Gong, Y.~Fu, and T.~M. Hospedales, ``A simple feature
  augmentation for domain generalization,'' in \emph{International Conference
  on Computer Vision (ICCV)}, 2021, pp. 8886--8895.

\bibitem{DBLP:conf/iclr/ZhouY0X21}
K.~Zhou, Y.~Yang, Y.~Qiao, and T.~Xiang, ``Domain generalization with
  mixstyle,'' in \emph{International Conference on Learning Representations
  (ICLR)}, 2021.

\bibitem{DBLP:conf/iccv/BustoG17}
P.~P. Busto and J.~Gall, ``Open set domain adaptation,'' in \emph{International
  Conference on Computer Vision (ICCV)}, 2017, pp. 754--763.

\bibitem{DBLP:conf/eccv/SaitoYUH18}
K.~Saito, S.~Yamamoto, Y.~Ushiku, and T.~Harada, ``Open set domain adaptation
  by backpropagation,'' in \emph{European Conference on Computer Vision
  (ECCV)}, 2018, pp. 156--171.

\bibitem{DBLP:conf/cvpr/LiuCLW019}
H.~Liu, Z.~Cao, M.~Long, J.~Wang, and Q.~Yang, ``Separate to adapt: Open set
  domain adaptation via progressive separation,'' in \emph{IEEE Conference on
  Computer Vision and Pattern Recognition (CVPR)}, 2019, pp. 2927--2936.

\bibitem{DBLP:conf/cvpr/PanYLNM20}
Y.~Pan, T.~Yao, Y.~Li, C.~Ngo, and T.~Mei, ``Exploring category-agnostic
  clusters for open-set domain adaptation,'' in \emph{IEEE Conference on
  Computer Vision and Pattern Recognition (CVPR)}, 2020, pp. 13\,864--13\,872.

\bibitem{DBLP:conf/icml/LuoWHB20}
Y.~Luo, Z.~Wang, Z.~Huang, and M.~Baktashmotlagh, ``Progressive graph learning
  for open-set domain adaptation,'' in \emph{International Conference on
  Machine Learning (ICML)}, 2020, pp. 6468--6478.

\bibitem{DBLP:journals/tmm/SherminLTMS21}
T.~Shermin, G.~Lu, S.~W. Teng, M.~M. Murshed, and F.~Sohel, ``Adversarial
  network with multiple classifiers for open set domain adaptation,''
  \emph{IEEE Transactions on Multimedia (TMM)}, vol.~23, pp. 2732--2744, 2021.

\bibitem{DBLP:journals/jmlr/GrettonBRSS12}
A.~Gretton, K.~M. Borgwardt, M.~J. Rasch, B.~Sch{\"{o}}lkopf, and A.~J. Smola,
  ``A kernel two-sample test,'' \emph{Journal of Machine Learning Research
  (JMLR)}, vol.~13, no.~1, pp. 723--773, 2012.

\bibitem{DBLP:journals/tmm/LuoJGLLLG20}
H.~Luo, W.~Jiang, Y.~Gu, F.~Liu, X.~Liao, S.~Lai, and J.~Gu, ``A strong
  baseline and batch normalization neck for deep person re-identification,''
  \emph{IEEE Transactions on Multimedia (TMM)}, vol.~22, no.~10, pp.
  2597--2609, 2020.

\bibitem{DBLP:conf/iccv/FuWWZSUH19}
Y.~Fu, Y.~Wei, G.~Wang, Y.~Zhou, H.~Shi, and T.~S. Huang, ``Self-similarity
  grouping: A simple unsupervised cross domain adaptation approach for person
  re-identification,'' in \emph{International Conference on Computer Vision
  (ICCV)}, 2019, pp. 6111--6120.

\bibitem{DBLP:conf/iccv/ZhengSTWWT15}
L.~Zheng, L.~Shen, L.~Tian, S.~Wang, J.~Wang, and Q.~Tian, ``Scalable person
  re-identification: {A} benchmark,'' in \emph{International Conference on
  Computer Vision (ICCV)}, 2015, pp. 1116--1124.

\bibitem{DBLP:conf/iccv/ZhengZY17}
Z.~Zheng, L.~Zheng, and Y.~Yang, ``Unlabeled samples generated by {GAN} improve
  the person re-identification baseline in vitro,'' in \emph{International
  Conference on Computer Vision (ICCV)}, 2017, pp. 3774--3782.

\bibitem{wei2018person}
L.~Wei, S.~Zhang, W.~Gao, and Q.~Tian, ``Person transfer gan to bridge domain
  gap for person re-identification,'' in \emph{IEEE Conference on Computer
  Vision and Pattern Recognition (CVPR)}, 2018, pp. 79--88.

\bibitem{DBLP:conf/cvpr/LiZXW14}
W.~Li, R.~Zhao, T.~Xiao, and X.~Wang, ``Deepreid: Deep filter pairing neural
  network for person re-identification,'' in \emph{IEEE Conference on Computer
  Vision and Pattern Recognition (CVPR)}, 2014, pp. 152--159.

\bibitem{DBLP:conf/cvpr/ZhongZCL17}
Z.~Zhong, L.~Zheng, D.~Cao, and S.~Li, ``Re-ranking person re-identification
  with k-reciprocal encoding,'' in \emph{IEEE Conference on Computer Vision and
  Pattern Recognition (CVPR)}, 2017, pp. 3652--3661.

\bibitem{DBLP:conf/cvpr/LiW13}
W.~Li and X.~Wang, ``Locally aligned feature transforms across views,'' in
  \emph{IEEE Conference on Computer Vision and Pattern Recognition (CVPR)},
  2013, pp. 3594--3601.

\bibitem{xiao2016end}
T.~Xiao, S.~Li, B.~Wang, L.~Lin, and X.~Wang, ``End-to-end deep learning for
  person search,'' \emph{arXiv preprint arXiv:1604.01850}, 2016.

\bibitem{DBLP:conf/scia/HirzerBRB11}
M.~Hirzer, C.~Beleznai, P.~M. Roth, and H.~Bischof, ``Person re-identification
  by descriptive and discriminative classification,'' in \emph{Scandinavian
  Conference on Image Analysis (SCIA)}, 2011, pp. 91--102.

\bibitem{DBLP:conf/eccv/GrayT08}
D.~Gray and H.~Tao, ``Viewpoint invariant pedestrian recognition with an
  ensemble of localized features,'' in \emph{European Conference on Computer
  Vision (ECCV)}, 2008, pp. 262--275.

\bibitem{DBLP:conf/cvpr/HeZRS16}
K.~He, X.~Zhang, S.~Ren, and J.~Sun, ``Deep residual learning for image
  recognition,'' in \emph{IEEE Conference on Computer Vision and Pattern
  Recognition (CVPR)}, 2016, pp. 770--778.

\bibitem{DBLP:conf/eccv/PanLST18}
X.~Pan, P.~Luo, J.~Shi, and X.~Tang, ``Two at once: Enhancing learning and
  generalization capacities via ibn-net,'' in \emph{European Conference on
  Computer Vision (ECCV)}, 2018, pp. 484--500.

\bibitem{DBLP:conf/cvpr/DengDSLL009}
J.~Deng, W.~Dong, R.~Socher, L.~Li, K.~Li, and F.~Li, ``Imagenet: {A}
  large-scale hierarchical image database,'' in \emph{IEEE Conference on
  Computer Vision and Pattern Recognition (CVPR)}, 2009, pp. 248--255.

\bibitem{cubuk2018autoaugment}
E.~D. Cubuk, B.~Zoph, D.~Mane, V.~Vasudevan, and Q.~V. Le, ``Autoaugment:
  Learning augmentation policies from data,'' \emph{arXiv preprint
  arXiv:1805.09501}, 2018.

\bibitem{zhang2017alignedreid}
X.~Zhang, H.~Luo, X.~Fan, W.~Xiang, Y.~Sun, Q.~Xiao, W.~Jiang, C.~Zhang, and
  J.~Sun, ``Alignedreid: Surpassing human-level performance in person
  re-identification,'' \emph{arXiv preprint arXiv:1711.08184}, 2017.

\bibitem{nichol2018first}
A.~Nichol, J.~Achiam, and J.~Schulman, ``On first-order meta-learning
  algorithms,'' \emph{arXiv preprint arXiv:1803.02999}, 2018.

\bibitem{DBLP:conf/iclr/ShankarPCCJS18}
S.~Shankar, V.~Piratla, S.~Chakrabarti, S.~Chaudhuri, P.~Jyothi, and
  S.~Sarawagi, ``Generalizing across domains via cross-gradient training,'' in
  \emph{International Conference on Learning Representations (ICLR)}, 2018.

\bibitem{DBLP:journals/pami/SunZLYTW21}
Y.~Sun, L.~Zheng, Y.~Li, Y.~Yang, Q.~Tian, and S.~Wang, ``Learning part-based
  convolutional features for person re-identification,'' \emph{IEEE
  Transactions on Pattern Analysis and Machine Intelligence}, vol.~43, no.~3,
  pp. 902--917, 2021.

\bibitem{DBLP:conf/aaai/LiYSH18}
D.~Li, Y.~Yang, Y.~Song, and T.~M. Hospedales, ``Learning to generalize:
  Meta-learning for domain generalization,'' in \emph{AAAI Conference on
  Artificial Intelligence (AAAI)}, 2018, pp. 3490--3497.

\bibitem{DBLP:conf/cvpr/QiaoLSY18}
S.~Qiao, C.~Liu, W.~Shen, and A.~L. Yuille, ``Few-shot image recognition by
  predicting parameters from activations,'' in \emph{IEEE Conference on
  Computer Vision and Pattern Recognition (CVPR)}, 2018, pp. 7229--7238.

\bibitem{DBLP:conf/eccv/WenZL016}
Y.~Wen, K.~Zhang, Z.~Li, and Y.~Qiao, ``A discriminative feature learning
  approach for deep face recognition,'' in \emph{European Conference on
  Computer Vision (ECCV)}, 2016, pp. 499--515.

\bibitem{van2008visualizing}
L.~Van~der Maaten and G.~Hinton, ``Visualizing data using t-sne,''
  \emph{Journal of machine learning research (JMLR)}, vol.~9, no.~11, pp.
  2579--2605, 2008.

\end{thebibliography}

\end{document}